\documentclass{article}

\usepackage{arxiv}
\usepackage{textcomp}

\usepackage[utf8]{inputenc} 
\usepackage[T1]{fontenc}    
\usepackage{enumerate}
\usepackage{hyperref}       
\usepackage{url}            
\usepackage{booktabs}       
\usepackage{amsfonts}       
\usepackage{amsmath}        %
\usepackage{gensymb}        %
\usepackage{nicefrac}       
\usepackage{microtype}      
\usepackage{lipsum}
\usepackage{graphicx}
\usepackage{wrapfig}
\usepackage{float}
\usepackage{todonotes}
\graphicspath{ {./images/} }
\usepackage{svg}

\title{On sparse connectivity, adversarial robustness, and a novel model of the artificial neuron}

\author{
 Sergey Bochkanov \\
 ALGLIB Project \\
 Russian Federation \\
  \texttt{sergey.bochkanov@alglib.net} \\
}

\begin{document}
\maketitle
\begin{abstract}
Deep neural networks have achieved human-level accuracy on almost all perceptual benchmarks.
It is interesting that these advances were made using two ideas that are decades old: (a) an artificial neuron based on a linear summator and (b) SGD training. 

However, there are important metrics beyond accuracy: computational efficiency and stability against adversarial perturbations.
In this paper, we propose two closely connected methods to improve these metrics on contour recognition tasks:
(a) a novel model of an artificial neuron, a "strong neuron," with low hardware requirements and inherent robustness against adversarial perturbations
and (b) a novel constructive training algorithm that generates sparse networks with $O(1)$ connections per neuron.

We demonstrate the feasibility of our approach through experiments on SVHN and GTSRB benchmarks.
We achieved an impressive 10x-100x reduction in operations count (10x when compared with other sparsification approaches, 100x when compared with dense networks) and a substantial reduction in hardware requirements (8-bit fixed-point math was used) with no reduction in model accuracy.
Superior stability against adversarial perturbations (exceeding that of adversarial training) was achieved without any counteradversarial measures, relying on the robustness of strong neurons alone.
We also proved that constituent blocks of our strong neuron are the only activation functions with perfect stability against adversarial attacks.
\end{abstract}


\section{Introduction}

In recent decades, artificial neural networks have achieved impressive results on all computer vision benchmarks.
Perhaps the correct phrase would be "unbelievably good" because a hypothetical time traveller from the year 2000 would be shocked by today's progress in this area.

One could have predicted, relying on Moore's law, the computing power of today's CPUs.
However, it would have been impossible to predict the completely unexpected success in the training of large nonconvex multiextremal models --- object recognition, neural text translation, style transfer, and deep fakes.
Interestingly, this progress was achieved using two ideas that are decades old: (1) an artificial neuron with a linear summator at its core and (2) stochastic gradient (SGD) training.

The combination of these ideas was fortuitous, allowing us to fit any decision function, no matter how complex.
As a result, in recent years neural models surpassed human-level accuracy on ImageNet and other benchmarks.
However, we believe (and will justify below) that the very properties of summators and SGD impede progress in improving two other important metrics: the sparsity of the neural connections and adversarial stability.

In our work, we propose (1) a novel model of an artificial neuron with inherent robustness against adversarial perturbations and (2) a novel training algorithm that allows us to build extremely sparse networks with $O(1)$ connections per neuron.
With these proposals, we achieved state-of-the-art performance and adversarial stability on a number of contour recognition benchmarks.

The article is structured as follows.
In section \ref{sect:novelneuron}, we will discuss the deficiencies of linear summators and propose a new model of an artificial neuron that we call the "strong neuron."
In section \ref{sect:rationale}, we will show that the structure of our strong neuron is motivated by obvious stability requirements and that our strong neuron is the only perfectly stable artificial neuron possible.
In section \ref{sect:overview}, we will discuss three blocks of the Contour Engine, a neural architecture that utilizes our proposed strong neurons: a feature detection unit, sparse inference unit, and shallow classifier.
The key part of our network --- the sparsely connected geometric inference engine --- and its training algorithm will be discussed in section \ref{sect:sparselayers}.
The initial feature detection layer will be briefly discussed in section \ref{sect:featuredetector} (with a more detailed discussion in Appendix B).
The shallow classifier that performs post-processing of the network output will be discussed in section \ref{sect:shallowclassifier}.
In section \ref{sect:comparison}, we will compare our architecture with similar and related approaches.
In section \ref{sect:results}, we will discuss the experimental results.
Finally, in section \ref{sect:conclusions}, we present a brief summary of our findings and a few thoughts on future research directions.

\section{The novel artificial neuron ("strong neuron")}
\label{sect:novelneuron}

In this work we propose to replace traditional summator-based artificial neurons with a more powerful one that (a) can separate input images with decision surfaces much more complex than hyperplanes, (b) has better stability properties with respect to the adversarial perturbations of its inputs, (c) inherently favors sparsity of connections and (d) has fairly low hardware requirements (8-bit fixed point hardware is enough in most cases).

\begin{figure}[h!]
    \centering
    \includegraphics[width=10cm]{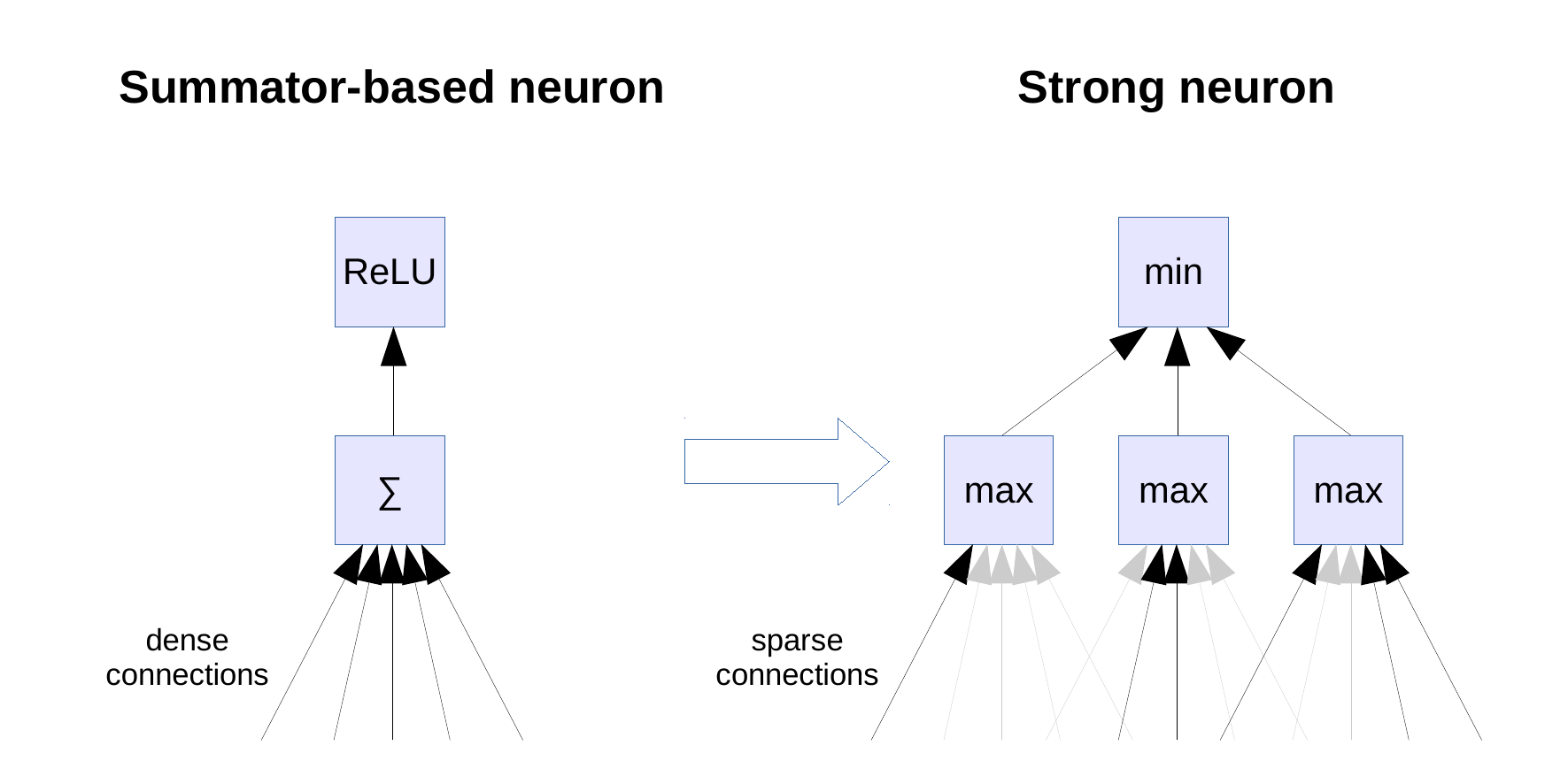}
    \caption{A summator-based neuron and a strong neuron}
    \label{fig:fig1_strongnn}
\end{figure}

In the following subsections, we discuss specifics of the contour recognition problems, strong and weak points of the summator-based artificial neuron and, finally, our proposal.

\subsection{Contour recognition = logical AND + logical OR}

Contour recognition is an important subset of computer vision problems.
It is deeply connected with properties of our world --- we live in a universe full of localized objects with distinctive edges.
Many important problems are contour based: handwritten digit recognition, traffic light detection, traffic sign recognition and number plate recognition.

There are also non-contour tasks --- for example, ones that can only be solved by gathering information from many small cues scattered throughout an image (e.g., distinguishing a food store from an electronics store).
A degenerate counterexample is a task that involves computing the mean intensity of the image pixels --- its decision function ignores any kind of spatial structure in the image.

Contour recognition has interesting mathematical properties:

\begin{itemize}
\item
It naturally leads to $[0,1]$-bounded activities.
Not all computer vision problems have this property (e.g., object counting tasks have unbounded activities).
\item
Contours are localized and independent from their surrounding (e.g., a crosswalk sign is a crosswalk sign, regardless of who uses the crosswalk --- a pedestrian, a tank or a bird).
\item
Ideal contour detector should have a monotonic response with respect to the full/partial "dimming" of the contour or some of its parts.
In other words, if you start to progressively remove parts of the contour, you should observe monotonically decreasing detector responses.
\end{itemize}

Our insight is that contour recognition is essentially a combination of two basic operations on low-level features:

\begin{itemize}
\item logical AND (detection), which decomposes high-level features as combinations of several low-level ones, placed at different locations
\item logical OR (generalization), which allows detectors to be activated by more diverse inputs
\end{itemize}

\begin{figure}[H]
    \centering
    \includegraphics[width=10cm]{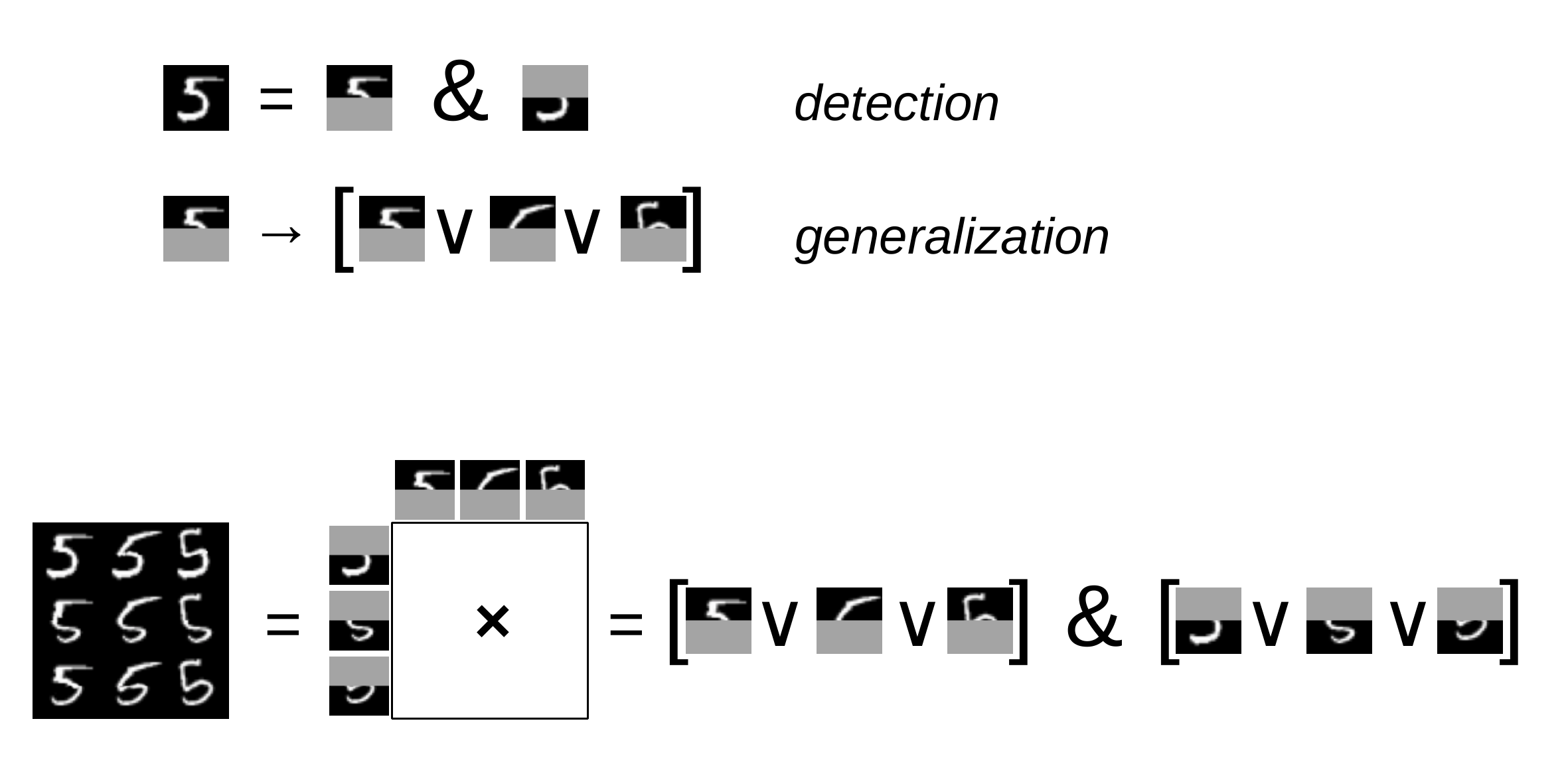}
    \caption{Pattern recognition: AND + OR}
    \label{fig:andor}
\end{figure}

\subsection{What is wrong with linear summator and SGD?}

A linear summator trained with SGD is an excellent basic building block for a number of reasons:

\begin{itemize}
\item 
First, it is flexible.
It smoothly implements soft-AND/soft-OR logic within a single framework: $AND_{RELU}(A,B)=ReLU(A+B-1)$, $OR_{RELU}(A,B)=ReLU(A+B)$.
It may also implement more general decision functions (including ones with negative weights).
\item 
Second, it is trainable.
We usually accept it as a given that one can stack many linear units interleaved with nonlinearities, constructe a huge nonlinear nonconvex model and \emph{successfully} fit it with SGD to some complex and noisy decision function.
\end{itemize}

However, it has some deficiencies as well
First, summator-based implementation of the AND/OR logic is very brittle, especially in high-dimensional spaces.
The neuron can be set to an arbitrarily high value (or, alternatively, zeroed) by feeding it with many small activities in different channels.
Many researchers believe that this is the reason behind the adversarial instability of modern neural networks.

We also feel (more intuition that concrete proof) that SGD-based training has limited potential for sparsification.
There are multiple sparsification strategies that share one common trait: they start from the same dense network and progressively sparsify it (via $L_1$ regularization or by other means).
As a result, the final connection count is typically \emph{a fraction} of the initial connection count: $O(s{\times}C)$, where $s$ is a sparsity coefficient that may be quite small --- 0.1, 0.01 or even less --- although it is asymptotically different from zero.
Thus, we believe that sparsity via regularization is inferior to sparsity achieved by other means (explicit channel selection or sparsifying constraints).

\subsection{Our proposal}

\begin{figure}[ht]
    \centering
    \includegraphics[width=12cm]{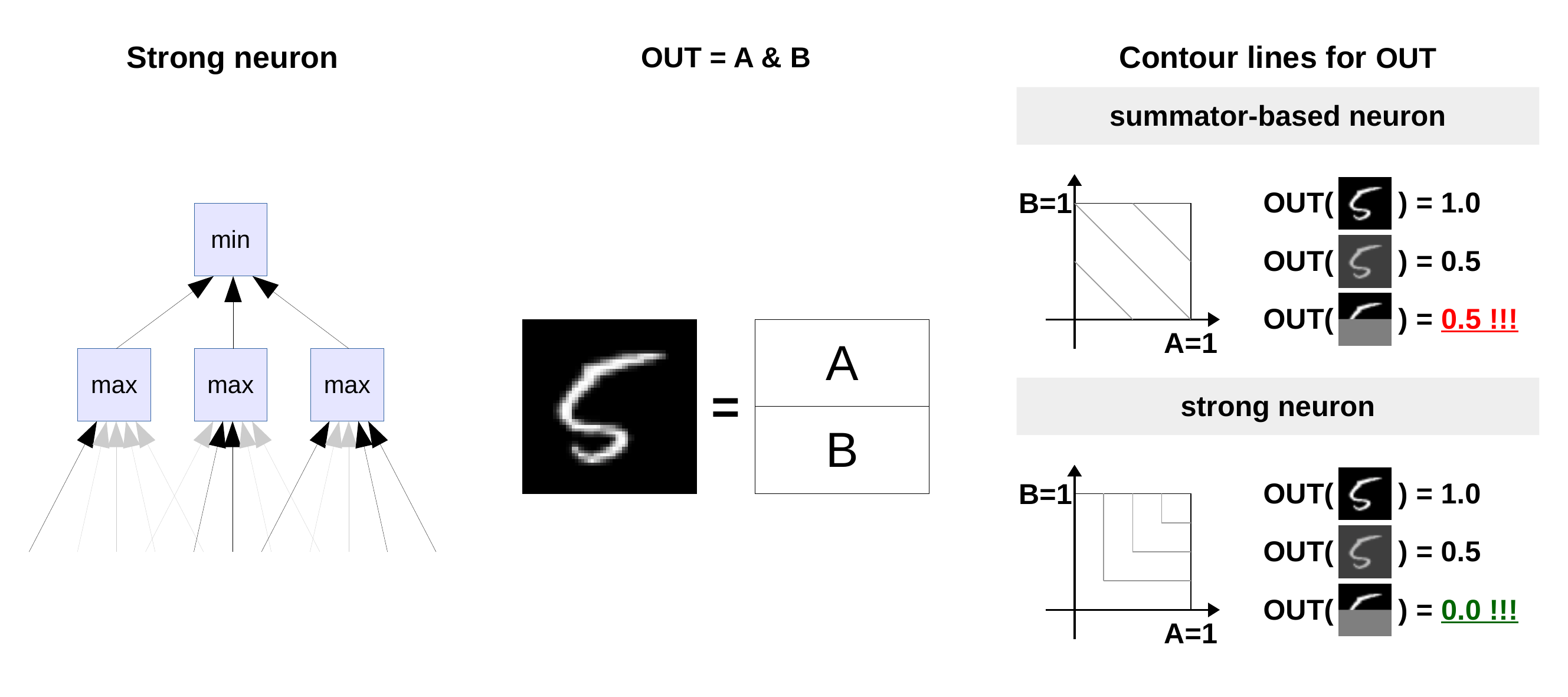}
    \caption{The strong neuron is better at pattern recognition than the linear one}
    \label{fig:stronger}
\end{figure}

We propose to use f(A,B)=min(A,B,1) to implement AND-logic, to use f(A,B)=max(A,B,0) to implement OR-logic and to combine both kinds of logic in a novel summator-free artificial neuron --- "strong neuron" (see Figure \ref{fig:stronger}).

We call our artificial neuron "strong" because it has a much more complex decision boundary than the summator-based neuron.
The shape of this boundary naturally fits into the pattern recognition framework.
Even with binary weights (which allowed us to achieve state-of-the-art results on GTSRB and SVHN benchmarks), standalone strong neurons can separate large chunks of the target class from the rest of the training set.

In the somewhat exaggerated example shown in Figure \ref{fig:stronger}, the standalone summator-based neuron cannot distinguish between the full image dimmed by 50\% (reduced contrast) and the image with a completely dropped bottom half.
The linearity of the summator means that it is possible to compensate for the lack of activity in one channel by increasing the activity in another one.
In contrast, the strong neuron easily and naturally distinguishes between these two images.

Another important property of our strong neuron is that its amplification of adversarial perturbations can be precisely controlled.
Further, with binary weights the layer of strong neurons becomes robust with respect to adversarial attacks: an $\epsilon$-bounded perturbation of inputs produces exactly $\epsilon$-bounded perturbation of outputs.

We also propose a novel training algorithm that can train strong neurons with sparse connectivity.
This algorithm reformulates the initial nonlinear least squares problem subject to sparsity constraints as a discrete one problem with discrete (binary or nonbinary) weights and discrete sparsity constraints, which is efficiently solved by the newly proposed heuristic.

The properties of strong neurons and their training algorithm can be used to reduce hardware requirements --- in particular, to avoid expensive floating point units.
With binary weights, our strong neurons are summation-free and multiplication-free --- only $min$ and $max$ operations are needed to implement strong neurons.
Moreover, the adversarial stability of strong neurons means that they are also resistant to random perturbations from rounding errors (i.e., it is possible to reduce precision from full 32-bit floating point to 8-bit fixed-point without sacrificing inference accuracy).

\section{The motivation behind our model}
\label{sect:rationale}

In this section, we will show that our artificial neuron model is motivated by some fundamental considerations, that is, there are some reasonable and intuitive requirements that are satisfied by our model --- and are not satisfied by summator-based neurons.

First, we define the $L_\infty$-nonexpansive function as one which in a general N-dimensional case satisfies

\begin{align*}
|f(x+{\Delta}x)-f(x)| \leq \max\limits_i|{\Delta}x_i| = {\lVert}{\Delta}x{\rVert}_\infty
\end{align*}

for any N-dimensional input perturbation ${\Delta}x$.
Similarly, we define the $L_1$-nonexpansive function as one that satisfies

\begin{align*}
|f(x+{\Delta}x)-f(x)| \leq \sum\limits_i|{\Delta}x_i| = {\lVert}{\Delta}x{\rVert}_1
\end{align*}

Clearly, both kinds of nonexpansive functions produce bounded output under bounded input perturbation.
However, the $L_\infty$ version provides stricter bounds than the $L_1$ one --- it does not accumulate perturbations.
For a 32x32x1 input image, $L_\infty$-nonexpansivity means that a change of $0.01$ in every pixel changes the output by at most $0.01$, and $L_1$-nonexpansivity means that the output change may be as large as $10.24=1024\times0.01$!

Another interesting question is how different kinds of nonexpansivity perform in a multilayer setting.
It is easy to see that $L_\infty$-nonexpansivity is preserved under superposition: $f_\infty(f_\infty(x),\dots,f_\infty(x))$ still produces an $\epsilon$-bounded output under an $\epsilon$-bounded input.
Conversely, stacking $L_1$-nonexpansive functions does not preserve this property: given that $f_1(x)$ produces an $N\epsilon$-bounded output under an $\epsilon$-bounded input, $f_1(f_1(x),\dots,f_1(x))$ will produce an $N^{2}\epsilon$-bounded output.

Human vision --- and any artificial vision system that should be robust --- has a bounded reaction to bounded perturbations of the input image.
The bounding ratio is not always 1:1 because sometimes we want to amplify weak signals.
Thus, enforcing $L_\infty$-nonexpansivity on the entire classifier may overconstrain it.
However, it makes sense to enforce this constraint at least for some parts of the classifier.
Our computational results show that stacking nonexpansive layers and performing potentially nonrobust inference only in the last step greatly improves stability against adversarial perturbations.

The rationale behind our model of the artificial neuron should be obvious --- making inference as robust as possible.
However, we present an even more interesting result --- the fact that our model is the only perfectly stable artificial neuron that implements AND/OR logic.

One familiar with the history of artificial neural networks may remember the so-called "XOR problem" --- a problem of fitting the simple four-point dataset below:

\begin{center}
\begin{tabular}[H]{ c c c }
 $x_0$ & $x_1$ & $y$ \\ 
 \hline
 0 & 0 & 0 \\
 0 & 1 & 1 \\
 1 & 0 & 1 \\
 1 & 1 & 0
\end{tabular}
\end{center}

This problem is an elegant example of a dataset that cannot be separated by the single linear summator.
Inspired by its minimalistic beauty, we formulate two similar problems, which address the accumulation of perturbations in multilayer networks:

\paragraph{Theorem 1: $L_\infty$-nonexpansive AND problem.}
$\exists!{\enspace}f(x,y)=min(x,y)$ such that the following holds:
\begin{enumerate}
\item $f(x,y)$ is defined for $x,y \in [0,1]$
\item $f(0,0)=f(0,1)=f(1,0)=0$
\item $f(1,1)=1$
\item $a{\leq}A,\ \ b{\leq}B \implies f(a,b){\leq}f(A,B)$ (monotonicity)
\item $|f(a+{\Delta}a,b+{\Delta}b)-f(a,b)| \leq max(|{\Delta}a|,|{\Delta}b|)$
\end{enumerate}

\paragraph{Theorem 2: $L_\infty$-nonexpansive OR problem.}
$\exists!{\enspace}g(x,y)=max(x,y)$ such that the following holds:
\begin{enumerate}
\item $g(x,y)$ is defined for $x,y \in [0,1]$
\item $g(0,0)=0$
\item $g(0,1)=g(1,0)=g(1,1)=1$
\item $a{\leq}A,\ \ b{\leq}B \implies g(a,b){\leq}g(A,B)$ (monotonicity)
\item $|g(a+{\Delta}a,b+{\Delta}b)-g(a,b)| \leq max(|{\Delta}a|,|{\Delta}b|)$
\end{enumerate}

Proofs of theorems 1 and 2 can be found in Appendix A \ref{sect:appendixa}.

These theorems have the following consequences:
\begin{itemize}
\item Our $min$-based AND and $max$-based OR elements are the only perfectly robust implementations of AND/OR logic
\item It is impossible to implement a robust AND (robust OR) element with just one ReLU neuron --- the best that can be achieved is $L_1$-nonexpansivity, which is not robust
\item It is possible to implement robust AND/OR logic by performing tricks with many traditional ReLU neurons ($max(a,b)=a+ReLU(b-a)$, $max(a,b,c)=max(a,max(b,c))$ and so on), but the result will be just another implementation of our robust AND/OR logic --- although it is much harder to achieve with SGD training
\end{itemize}

\section{Contour Engine: architecture overview}
\label{sect:overview}

In previous sections, we presented our model of the artificial neuron and discussed the motivation behind it, its significance and differences between the novel neuron and traditional summator-based ones.
In this section, we briefly discuss the architecture of our network before moving to more detailed explanations in the following sections.

\begin{figure}[H]
    \centering
    \includegraphics[width=14cm]{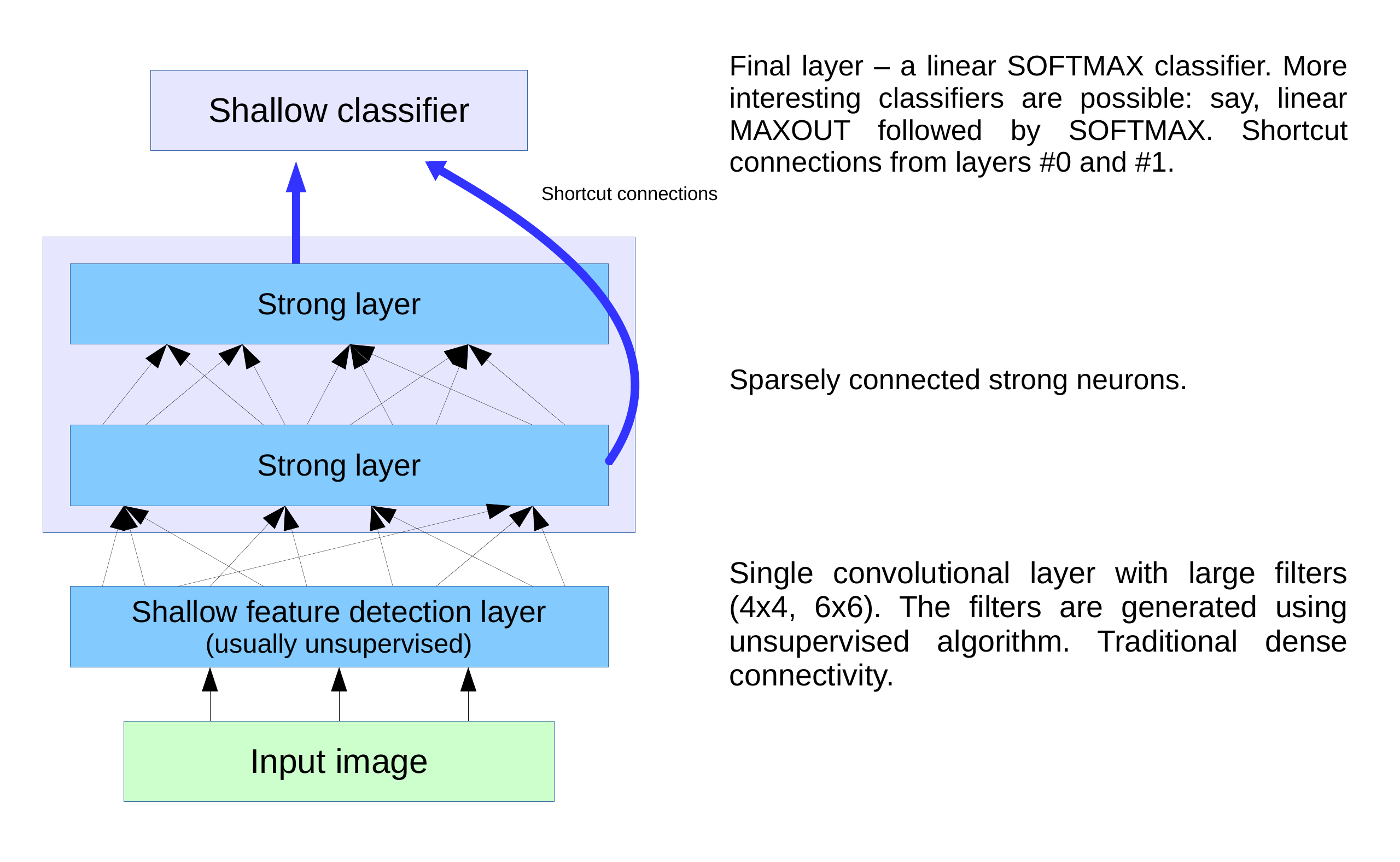}
    \caption{Three blocks of the Contour Engine network}
    \label{fig:contourengine}
\end{figure}

The three key parts of our neural architectures are:
\begin{itemize}
\item shallow feature detector
\item sparse contour detection layers
\item shallow classifier
\end{itemize}

The feature detection layer produces initial low-level features.
The contour detection layers (one or two is usually enough) combine them in order to produce medium and high-level features.
Finally, a linear or nonlinear classifier post-processes the features produced by the robust contour detection stage.

The training algorithm includes three distinct, sequential stages:

\begin{itemize}
\item train (iteratively) or build (noniteratively) a shallow feature detector
\item create sparse contour detection layers in a constructive manner (add layer by layer, create each layer neuron by neuron)
\item train a shallow classifier using activities of sparse layers as inputs
\end{itemize}

In our experiments, we used noniterative construction of the shallow feature detector --- either analytically constructed edge detection filters or filters obtained via unsupervised training were used (running k-means over image patches \cite{Coates11}).
Such an approach makes the input layer independent from label assignment, which allows us to make some interesting conclusions regarding the asymptotic complexity of the image recognition.

Our approach to the construction of sparse layers --- adding layers and neurons one by one --- is similar to and was inspired by the Cascade-Correlation network \cite{Fahlman90}.
The difference from the original work is that in order to generate new neurons we have to solve the \emph{nonsmooth} nonlinear least squares subproblem with additional sparsity $L_0$ constraints (for comparison, traditional summator-based neurons result in smooth unconstrained nonlinear least squares subproblems).

The second important contribution of our work (in addition to the robust artificial neuron) is the heuristic, which can efficiently find approximate solutions of such subproblems.
This heuristic is discussed in more detail in the next section.

Finally, the shallow classifier can be implemented as a linear layer (with SOFTMAX normalization) processing outputs of the sparse block.

\section{Training sparsely connected layers}
\label{sect:sparselayers}

This section discusses the core contribution of our work --- the constructive training of sparsely connected strong neurons.

\subsection{Issues with SGD training}

Based on our experience, online SGD training does not work well for networks with $min$-based activation functions.
We failed to achieve good results with SGD --- but maybe someone else will be able to do better.
We believe that the extreme nonconvexity of the $min$ function contributed to this failure ($max$ is less of a problem in our opinion), as it makes training much more difficult and prone to stalling in bad local extrema.

Our solution to these problems is the constructive training algorithm, which creates networks layer by layer, and each layer is created by adding neurons one by one.
This approach was investigated many times by many researchers with mixed results.
We again refer here to the work of Fahlman et al. on the Cascade-Correlation network \cite{Fahlman90}, which, in our opinion, was the most successful one and inspired our own research.

\subsection{The constructive training algorithm}

Training networks composed of highly nonconvex and nonsmooth elements is difficult.
Suppose, however, that \emph{somehow} you can train just one such element to fit some target function of your choice.
How can it help you train a network?
The answer is to build your model incrementally, training new elements to fit the current residual and adding them one by one.

\begin{figure}[H]
    \centering
    \includegraphics[width=14cm]{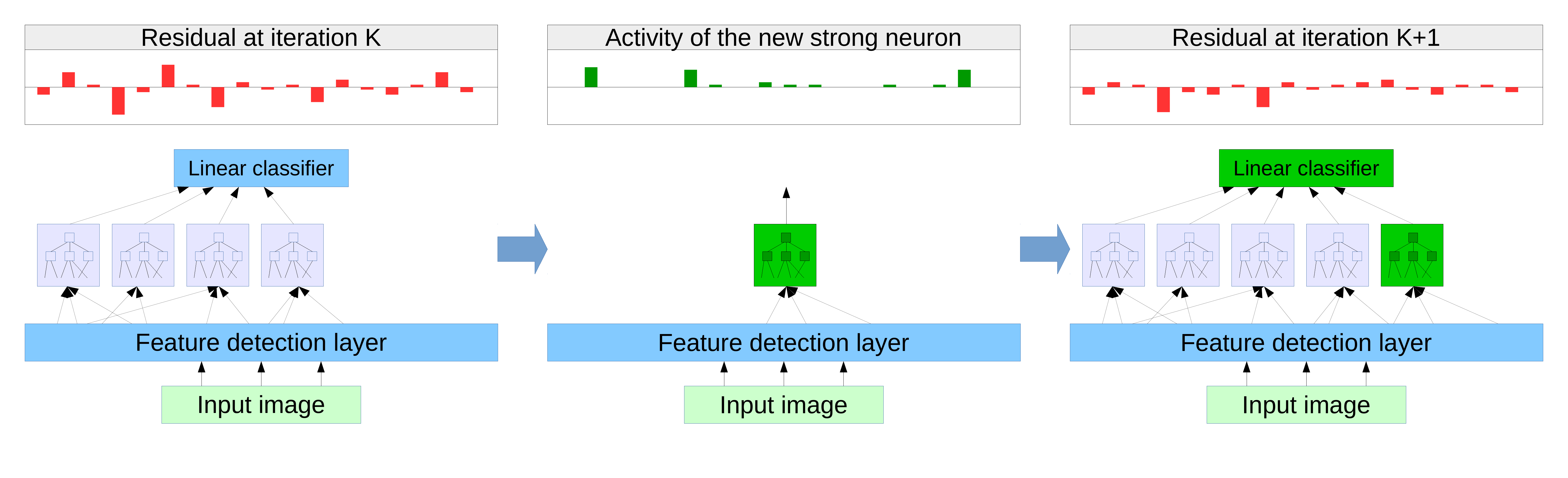}
    \caption{Incremental training procedure}
    \label{fig:trainlayers}
\end{figure}

New neurons are trained to fit the current residual of the classifier, and every time you add a neuron to the layer you have to retrain the classifier to obtain new residuals.
One may see some similarity to boosting here (we will return to this point later).

The algorithm listed above can be easily generalized to multilayer training.
One choice to be made is whether or not to maintain shortcut connections to the classifier from the previously learned layer.
The training procedure can easily fast-forward information from bottom to top by learning identity mapping if necessary, so it is mostly a matter of taste.

\subsection{Training strong neurons}

In the subsection above, we reduced the problem of training sparse multilayer networks to training just one neuron with sparse connections:

\begin{align*} 
\min\limits_{w} \sum\limits_{i}\left(N(w,X_i)-y_i\right)^2\ \ \ s.t.\ \ sparsity\ \ constraints
\end{align*}

where $w$ is a weight vector, $X_i$ is an $i$-th row of the input activities matrix $X$ (activities of the bottom layer at $i$-th image), $N(w,x)$ is a neuron output and $y_i$ is a target to fit (in our case, the current residual).

For a three-input strong neuron, the formulation above becomes:

\begin{equation} \label{eq:strong_nls_nonsmooth}
\begin{split}
\min\limits_{w_0, w_1, w_2} &\sum\limits_{i}\left[\min\left(\max\limits_{j}(w_{0,j}{\cdot}X_{i,j})\ ,\ \max\limits_{j}(w_{1,j}{\cdot}X_{i,j})\ ,\ \max\limits_{j}(w_{2,j}{\cdot}X_{i,j})\ ,\ \textbf{1}\right)-y_i\right]^2 s.t. \\
&{\lVert}w_0{\rVert}_0 \leq k\ ,\ \ {\lVert}w_1{\rVert}_0 \leq k\ ,\ \ {\lVert}w_2{\rVert}_0 \leq k
\end{split}
\end{equation}

This problem has no easy solution, even in an unconstrained setting, and $L_0$ constraints are hard to handle with present nonsmooth solvers.
Our proposal is to replace (\ref{eq:strong_nls_nonsmooth}) with some similar, albeit nonequivalent, form, which can be solved more efficiently and robustly.

One attractive property of the contour recognition problems is that they deal with $[0,1]$-bounded activities, where $0$ stands for the absence of some feature and $1$ stands for the maximum activity possible.
Thus, one may reasonably expect that all weights in (\ref{eq:strong_nls_nonsmooth}) will be nonnegative (connections with negative weights simply will not activate the neuron).
Furthermore, it makes sense to place further restrictions on the weights --- that is, to choose weights from some short fixed list, for example $\{0,\nicefrac{1}{2},1,1\nicefrac{1}{2},2\}$.

Now, instead of a nonconvex, nonsmooth, nonlinear least squares problem we have a combinatorial optimization problem:

\begin{equation} \label{eq:strong_nls_discrete}
\begin{split}
\min\limits_{w_0, w_1, w_2} &\sum\limits_{i}\left[\min\left(\max\limits_{j}(w_{0,j}{\cdot}X_{i,j})\ ,\ \max\limits_{j}(w_{1,j}{\cdot}X_{i,j})\ ,\ \max\limits_{j}(w_{2,j}{\cdot}X_{i,j})\ ,\ \textbf{1}\right)-y_i\right]^2 s.t. \\
&w_{0,j},w_{1,j},w_{2,j} \in W\\
&{\lVert}w_0{\rVert}_0 \leq k\ ,\ \ {\lVert}w_1{\rVert}_0 \leq k\ ,\ \ {\lVert}w_2{\rVert}_0 \leq k
\end{split}
\end{equation}

where $W$ can be binary $\{0,\ 1\}$ or something more fine-grained, such as $\{0,\ \nicefrac{1}{2},\ 1,\ 1\nicefrac{1}{2},\ 2\}$ or $\{0,\ \nicefrac{1}{4},\ \nicefrac{1}{2},\ \nicefrac{3}{4},\ 1,\ 1\nicefrac{1}{4},\ 1\nicefrac{1}{2},\ 1\nicefrac{3}{4},\ 2\}$.

Discrete optimization problems are usually harder to solve precisely than continuous ones.
Furthermore, \emph{this} discrete problem cannot be reduced to well-studied mixed-integer LP or mixed-integer QP, so there is likely no other way to solve it except for a brute-force search.
However, we do not need an exact solution --- having a good one is sufficient.
Our insight is that there is a simple heuristic that can generate good strong neurons without dealing with nonconvex multiextremal optimization problems.

The original discrete optimization problem has no constraints except for sparsity.
A $max$-element can gather information from any element of the input tensor (see figure below).
As a result, we have to evaluate prohibitively large amount of possible connection structures.
For instance, for 15 unit-weight connections to elements with a 32x32x20 input tensor we have roughly $10^{58}$ possible geometries.

\begin{figure}[H]
    \centering
    \includegraphics[width=8cm]{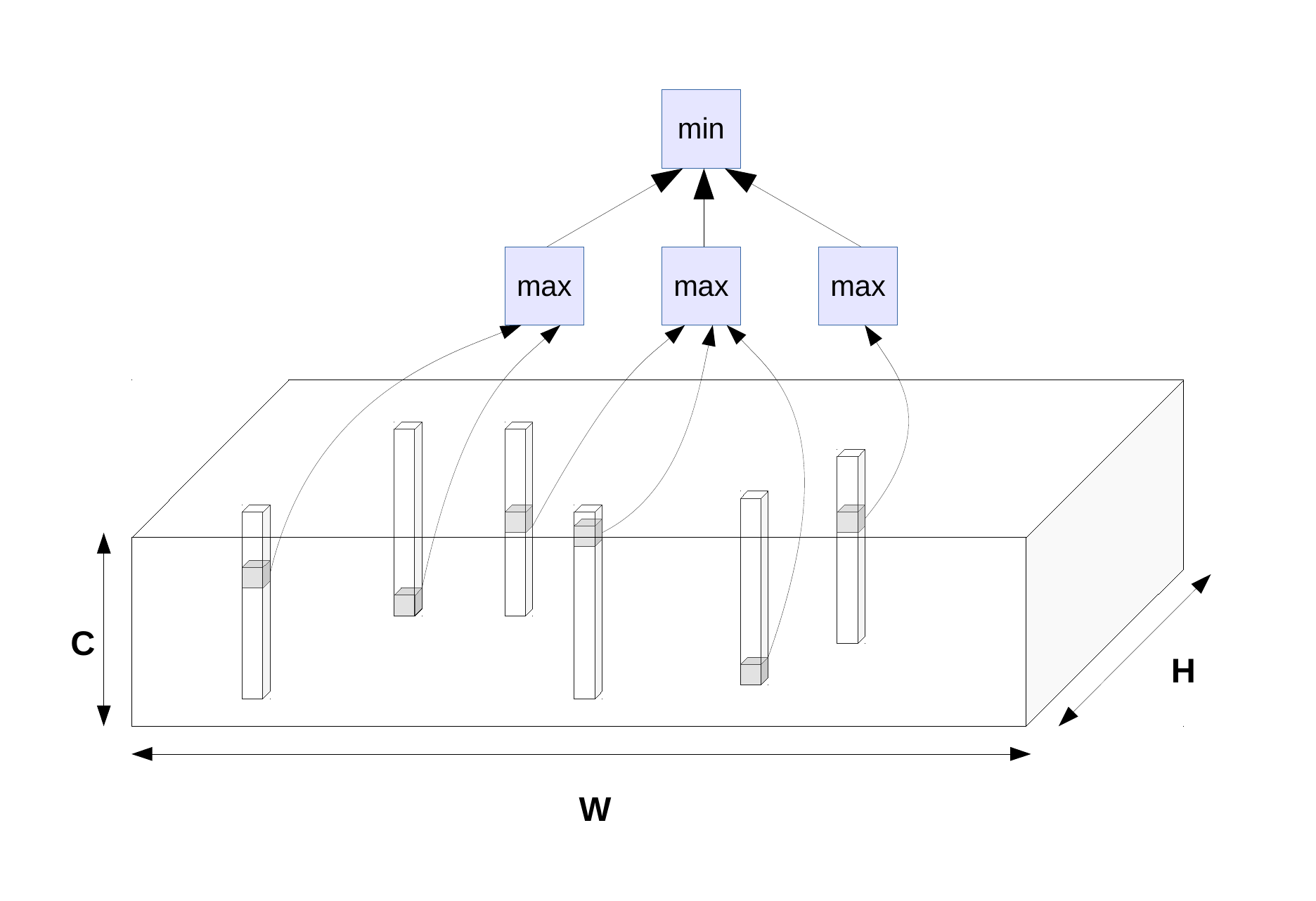}
    \caption{Totally unconstrained neuron}
    \label{fig:trn0}
\end{figure}

It is possible to significantly reduce the configuration count by adding some additional restrictions on the inter-layer connections.
For example, we may impose two additional constraints:
\begin{itemize}
\item Require that $max$-elements are spatially local (i.e., each element gathers inputs from just one location $(x,y)$ of the input tensor)
\item Require that $max$-elements feeding data into the same $min$-element are
located close to each other
\end{itemize}

Alternatively  --- for 1x1xD input tensors with no spatial component --- these restrictions can be reformulated as follows:
\begin{itemize}
\item Require that $max$-elements are correlationally local (i.e., each element gathers inputs from strongly correlated channels)
\item Require that $max$-elements feeding data into the same $min$-element are
correlated strongly enough
\end{itemize}

Having such constraints on the connections of the strong neuron significantly reduces the number of configurations that must be evaluated to solve the problem (\ref{eq:strong_nls_discrete}).
In our toy example, the configuration count is reduced from $10^{58}$ to just $10^{18}$.

\begin{figure}[H]
    \centering
    \includegraphics[width=8cm]{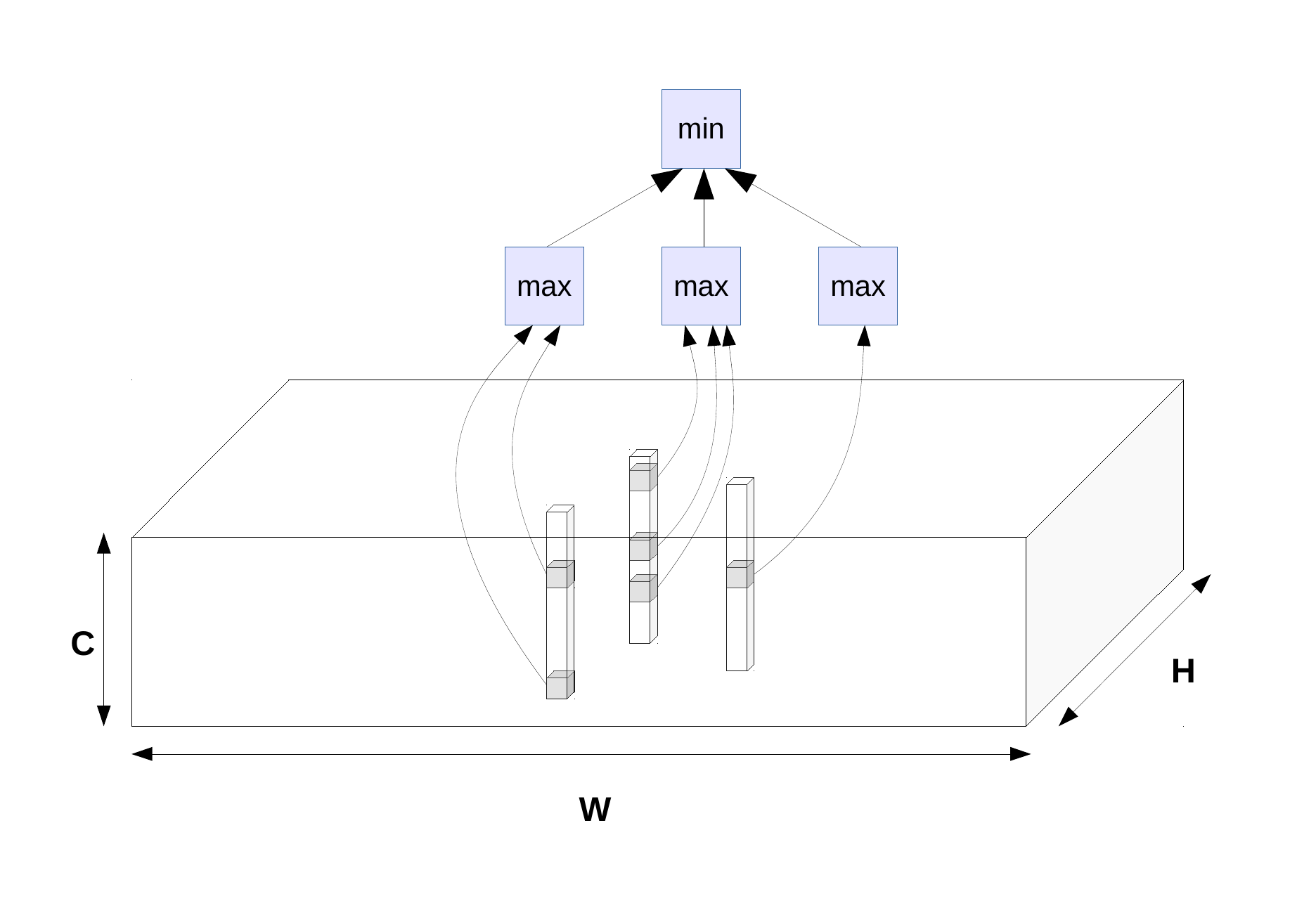}
    \caption{Strong neuron with spatial/correlational constraints}
    \label{fig:trn1}
\end{figure}

We can achieve a further reduction in search complexity through a two-step search procedure:
\begin{itemize}
\item Evaluate all possible "seed detectors" --- strong neurons with single-input $max$-elements (AND without OR)
\item Expand the best seed found --- sequentially add connections to its $max$-elements
\end{itemize}

\begin{figure}[H]
    \centering
    \includegraphics[width=8cm]{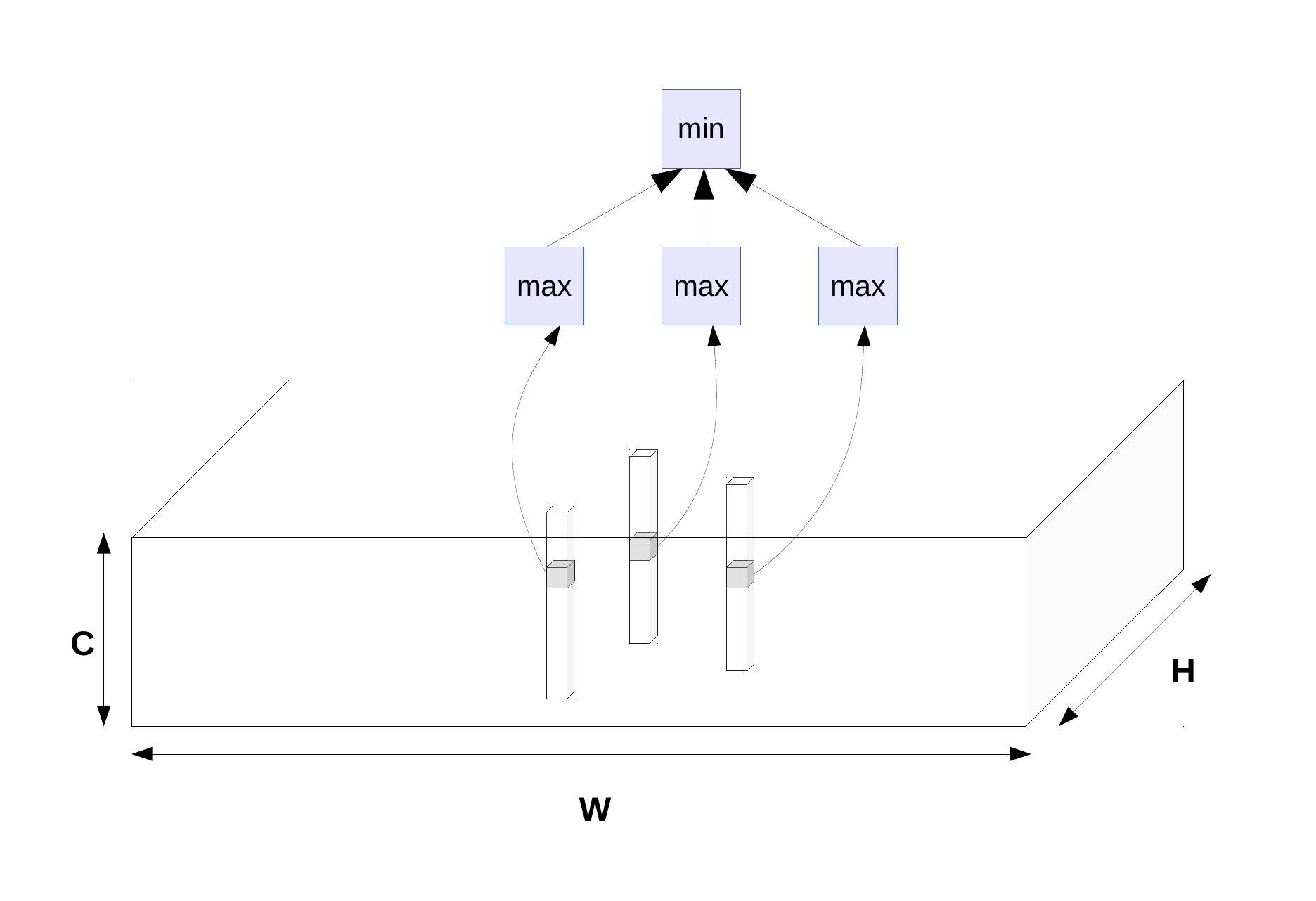}
    \caption{Seed detector --- a strong neuron without $max$-elements}
    \label{fig:trn2}
\end{figure}

As a result of this improvement, the search complexity for our 32x32x20 example is reduced from $10^{18}$ to $10^{9}$ neural configurations.
However, it is still too costly --- each of these configurations requires a full pass over the entire dataset in order to evaluate the neuron's performance.

Further improvements can be achieved by assuming the following:
\begin{itemize}
\item Good $f_3=\min(A,B,C)$ can be found by extending good $f_2=\min(A,B)$ with the best-suited $C$
\item Good $f_2=\min(A,B)$ can be found by extending good $f_1=A$ with the best-suited $B$
\item Good $f_1=A$ can be found by simply evaluating all possible single-input seed detectors
\end{itemize}

\begin{figure}[H]
    \centering
    \includegraphics[width=8cm]{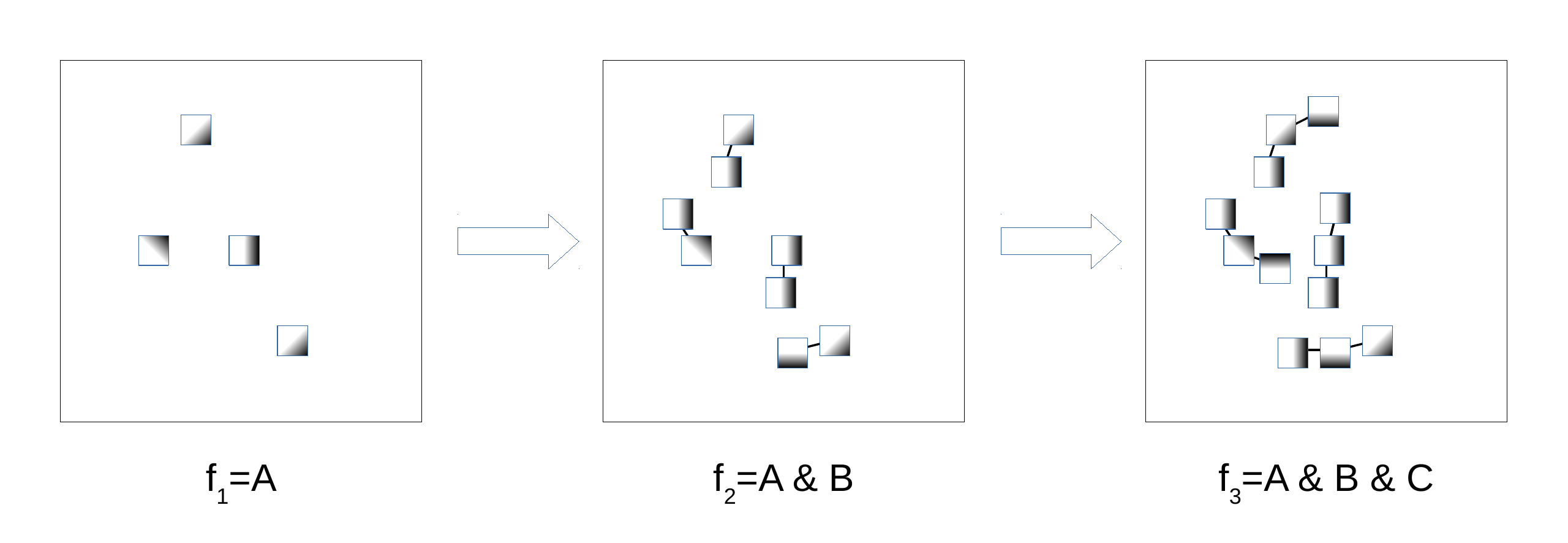}
    \caption{Growth of seed detectors}
    \label{fig:trn3}
\end{figure}

This improvement makes the problem (\ref{eq:strong_nls_discrete}) computationally tractable.
For example, the complexity of our toy example is reduced to just $20000$ combinations (compare this with the initial $10^{58}$ estimate).

\paragraph{Algorithm outline.} The simplified algorithm (only $\{0,1\}$ weights, input activities are $[0,1]$-bounded) is shown below:

\begin{enumerate}
\item Setup the initial model (empty with zero output) and a vector of its residuals over the entire dataset. Select a neuron pool size $P$ (a few hundreds works in most cases).
\item Competition phase: generate seed detectors and select the winner from the combined pool:
\begin{itemize}
\item Select a set of $P$ promising input features, "gen-1 seeds," $f_1=A$. Some form of quick and dirty feature selection is usually enough.
\item Produce $P$ gen-2 seeds by extending gen-1 seeds $f_1=A$ with such $B$ that $f_2=\min(A,B)$ produces the best linear fit to the current residual. Only the spatial/correlational neighborhood of $f_1$ is evaluated.
\item Produce $P$ gen-3 seeds by extending gen-2 seeds $f_2=\min(A,B)$ with such $C$ that $f_3=\min(A,B,C)$ produces the best linear fit to the current residual. Only the spatial/correlational neighborhood of $f_1$ is evaluated.
\end{itemize}
\item Generalization phase. Having determined a winning seed detector, sequentially extend its inputs with new $max$-connections:
\begin{itemize}
\item $f = \min(A, B, ...)$
\item $A \xrightarrow{} \max(A)$
\item $\max(A) \xrightarrow{} \max(A,A_2)$
\item $\max(A,A_2) \xrightarrow{} \max(A,A_2,A_3)$ and so on
\end{itemize}
Extending is performed in such a way that the extended detector fits the residual better than its previous version. Only the spatial/correlational neighborhood of $A$ is investigated. The procedure stops after the maximum number of connections is formed (good value --- 5 connections per $max$-element) or when there is no connection that can improve the fit.
\item Add a detector to the model, and update the classifier and residual vector. Stop after the user-specified amount of detectors is formed. Go to 2 otherwise.
\end{enumerate}

Although it is not explicitly stated, the algorithm above is a batch algorithm --- it requires us to keep an entire dataset in memory and make a full pass over it in order to generate new strong neurons.
The reason for this is that the algorithm has no way of correcting the neuron structure once it has been added to the model --- so, if you train a suboptimal neuron using a subsample of the entire training set, you will be unable to improve it later.
The only way to properly generate a neuron is to use all the available data.

This property raises an old question of the balance between network stability and its plasticity.
Networks trained with SGD have high plasticity but zero stability.
Plasticity allows us to use SGD --- an algorithm that makes only marginal improvements in the network being trained --- because these small decrements in the loss function will accumulate over time.
At the same time, it impedes cheap nondestructive retraining --- once an image is removed from the training set, it is quickly forgotten.

In contrast, our algorithm has zero plasticity --- it will not improve the neurons it generated previously --- but perfect stability.
The drawback of such an approach is that it is necessary to use an entire training set to generate just one strong neuron, and this job has to be done in the best way possible.
The upside is that the network never forgets what it learned before.
If your task has changed a bit, you can restart training and add a few new neurons without damaging previously learned ones.

\section{The feature detection layer}
\label{sect:featuredetector}

In this section, we briefly discuss the feature detection layer based on \cite{Coates11} and several proposed improvements.
We deem this part of our work as less important than the results discussed in the previous section (sparsely connected layers of the robust neurons).
Nevertheless, there are several interesting ideas we want to share here.
This section provides only a brief summary, with a detailed description presented in Appendix B \ref{sect:appendixb}.

\begin{wrapfigure}{r}{0.5\textwidth}
    \includegraphics[width=0.95\linewidth]{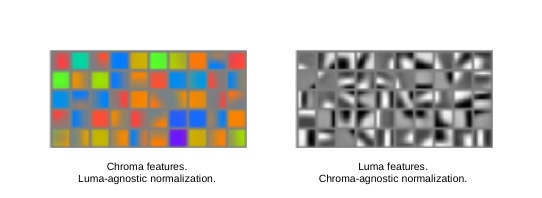}
    \caption{Filters learned with our (improved) procedure}
    \label{fig:chromaluma}
\end{wrapfigure}

Strong neurons can perform logical inference on low-level features, but they cannot \emph{produce} these features from raw pixel values.
Thus, a separate feature extraction block is essential in order to "prime" the Contour Engine.
The purpose of our feature extraction layer is to describe the input image using a rich dictionary of visual words.
The description includes features such as oriented edges, more complex shapes, colors and gradients, computed at multiple scales and orientations.

The key point of Coates et al. is that one may achieve surprisingly good classification performance by processing images with a single convolutional layer whose filters are trained in an unsupervised manner (k-means on random image patches).
The authors also proposed to post-process the raw convolutions with a simple activity sparsification filter $y_{sparse,i} = ReLU\left(y_i - \lambda\cdot mean(y)\right)$.

Filters as large as 4x4, 5x5 or 6x6 typically give the best results.
Figure \ref{fig:chromaluma} shows an example of the filters found with our training procedure.

We extend their results as follows:

\begin{itemize}
\item separate processing of color-agnostic (shape sensitive) and color-based features
\item multiple downsampling levels of the layer outputs (2x and 4x max-pooling are used together)
\item feature detection at multiple scales
\item completeness with respect to image transformations --- multiple versions of the same feature corresponding to positive/negative phases, permutations in color space, rotations and so on
\end{itemize}

\begin{figure}[H]
    \centering
    \includegraphics[width=10cm]{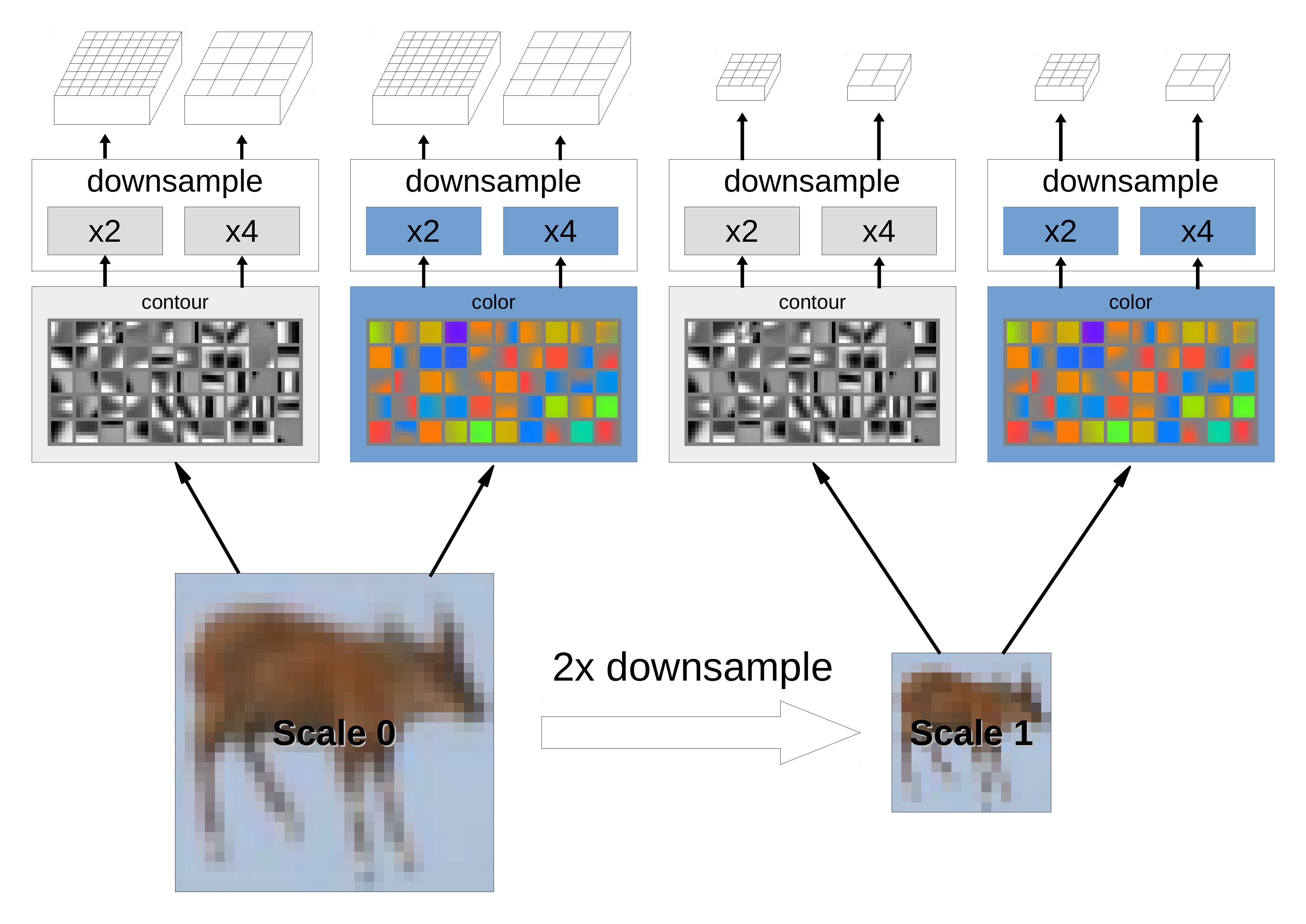}
    \caption{Multiscale multimodal feature extraction layer}
    \label{fig:v1layer}
\end{figure}

\section{The shallow classifier layer}
\label{sect:shallowclassifier}

Our proposed strong neurons have unique stability and sparsity properties, but some limitations are also present.
They have a rigid piecewise linear output with a fixed slope, but in order to separate image classes one often needs nonlinearities with steep slopes in some places and flat spots in other parts of the feature space.
Hence, a separate classifier layer is needed at the top of the network.

This classifier layer can be as deep as you wish --- but strong neurons perform data processing extremely well, so all you need in most cases is a single linear summator followed by SOFTMAX.
Training such a classifier is straightforward, requiring only sample activities of the bottom sparsely connected block over the entire dataset and training of the single-layer neural network (logit model) using the activities matrix as the input.

\emph{One important point to note is that the shallow classifier layer is the only place in our model where significant adversarial instability is introduced.}
The sparsely connected layers of strong neurons amplify adversarial perturbations in a completely controllable manner (and do not amplify them when binary weights are used).
The initial feature detection layer is a single layer of convolutions with bounded coefficients, and thus it has limited adversarial perturbation growth.

As a result, any adversary targeting our model will actually target its last layer.
In effect, this means that we reduced the problem of building a robust deep classifier to one of building a robust \emph{shalow} classifier.

In this work, we will show that, due to the stability of the bottom layers, a simple linear classifier performs well enough in terms of adversarial stability.

\section{Comparison with related approaches}
\label{sect:comparison}

In this section we discuss several other machine learning algorithms that are related to our work:

\begin{itemize}
\item Cascade-Correlation
\item Boosting
\item Forward-Thinking architecture
\item Deep neural decision forests
\item BagNet
\item $L_2$-nonexpansive networks
\end{itemize}

We also would like to briefly review some present defenses against adversarial attacks:

\begin{itemize}
\item Adversarial training
\item $L_2$-nonexpansive networks
\item Convex Outer Adversarial Polytope (Wong Defense)
\end{itemize}

\paragraph{Cascade-Correlation.}
We already mentioned and referred to the Cascade-Correlation architecture.
Our network construction algorithm reproduces Fahlman's idea in many respects.
Two important differences can be noted: (1) our algorithm trains sparsely connected strong neurons, and (2) unlike CasCor we try to avoid long chains of nonlinearities, which contribute to various instabilities, so our network has a shallow and wide layered structure.

\paragraph{Boosting.}
There is some similarity between our training algorithm and boosting.
Both algorithms expand the model by sequentially adding new units trained to fit the current residual.
Thus, one may consider our approach to be a special case of boosting.
However, boosting algorithms do not pay attention to the properties of weak classifiers added to the model; that is, any kind of weak classifier will fit into the boosting framework.
In contrast, robust strong neurons are essential to our network architecture.

\paragraph{Forward-Thinking architecture.}
Another interesting approach to discuss is Forward-Thinking architecture (see \cite{forwardthinking}).
This architecture is a constructive algorithm that trains the network layer by layer in a greedy manner.
Both Forward Thinking and Contour Engine use the same approach to create a layered network structure (different from both modern CNNs and Cascade-Correlation).

\paragraph{Deep neural decision forests.}
We also note some similarity between Contour Engine and one novel deep learning algorithm: deep neural decision forests \cite{deepneuraldf}.
First, there is a correspondence between our strong neurons and shallow decision trees.
Indeed, a strong neuron without $max$-units, the seed detector $f(A,B)=\min(A,B)$, is in some sense equivalent to a short decision tree.
One may generate such a tree, which returns $1$ for $A>0.5$ and $B>0.5$ and returns 0 otherwise.

The difference is that our strong neuron is more powerful than a shallow decision tree.
Adding $max$-connections achieves a quadratic/cubic increase in the model capacity with just a linear increase in its size.
Conversely, the capacity of the decision tree is linearly proportional to its size.

\paragraph{BagNet.}
BagNet, an experimental neural architecture \cite{bagnet}, achieves impressive classification results on ImageNet with the bag-of-local-features model.
By averaging predictions of the local models (each seeing just $\nicefrac{1}{7}\times\nicefrac{1}{7}$ of the entire image) it is possible to achieve results competitive with those of deep networks.

Authors have proposed this architecture as a proof of concept, which demonstrates that we have an incomplete understanding of the underlying mechanisms of computer vision algorithms.
For us, this approach is an interesting counterexample to Contour Engine.
Our architecture is based on a large-scale spatial structure, whereas BagNet works with scattered small-scale hints.

\paragraph{Adversarial training.}
A simple yet universal defense is to train the network using both original and adversarial examples\cite{advtrn}.
These additional examples make the inferences more robust by explicitly telling the network about the expected behavior under adversarial perturbation.
In theory, this may guide the network so that it will implement internally robust AND/OR logic (indeed, it is possible to implement $max$/$min$ with ReLU units).
The benefit of this approach is that it works for any kind of model --- all that is needed is a training code and a code that generates adversarial examples.

\paragraph{$L_2$-nonexpansive networks.}
This approach \cite{l2nonexpansive} is a class of neural networks in which "a unit amount of change in the inputs causes at most a unit amount of change in the outputs or any of the internal layers."
Due to the utilization of traditional summators, the authors were unable to achieve $L_\infty$-nonexpansivity, so they had to resort to weaker $L_2$-nonexpansivity (although it is still much better than $L_1$-nonexpansivity).

\paragraph{Convex Outer Adversarial Polytope (Wong Defense).}
This approach \cite{wongdefense} models network behavior under adversarial perturbation of its inputs.
An input image is provided along with per-component bounds of adversarial perturbation.
Wong's algorithm models the perturbation of activities of internal units and provides differentiable error bounds for network outputs.
It thus enables the use of straightforward SGD training on error bounds in order to reduce errors under adversarial perturbation.

\section{Experimental results}
\label{sect:results}

\subsection{Datasets}

We tested Contour Engine on two popular computer vision benchmarks: GTSRB and SVHN.

\paragraph{German Traffic Sign Recognition Benchmark.}
This benchmark is a multi-class single-image classification challenge \cite{gtsrb}.
The dataset has more than 50000 images of centered traffic signs belonging to 43 classes.
The classes are unequally sampled --- some "popular" traffic signs have many more instances than rare ones.
The images in the dataset were captured in the wild under slightly (sometimes wildly) different orientations, lighting conditions, image sizes (bounding rectangles from 18x18 pixels to 64x64 and larger) and amounts of motion blur.

\begin{figure}[H]
    \centering
    \includegraphics[width=5cm]{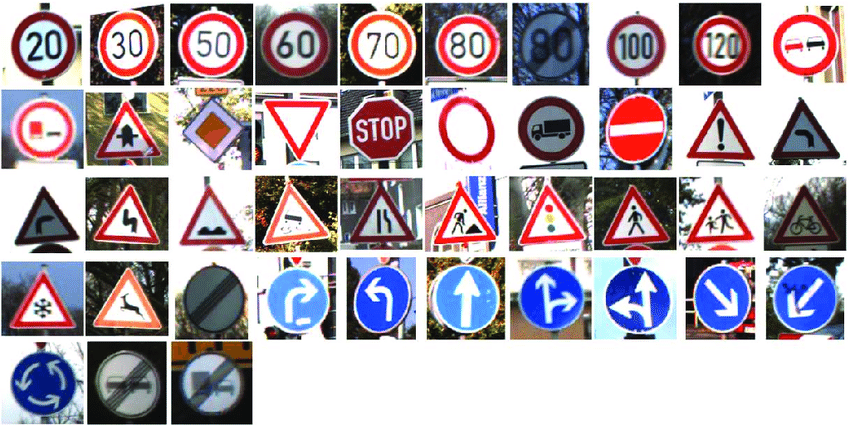}
    \caption{GTSRB dataset}
    \label{fig:gtsrb}
\end{figure}

We applied the following post-processing: we resized all images to standard 32x32 resolution, adding padding when necessary, and standardized brightness (mean 0.5).
In numerical experiments, affine distortions were used to augment the dataset.

\paragraph{Street View House Numbers.}
This dataset is a well-known 10-class digit recognition problem \cite{svhn}.
It has 630420 training and test images belonging to 10 classes.
The image size is 32x32 in all cases.

\begin{figure}[H]
    \centering
    \includegraphics[width=5cm]{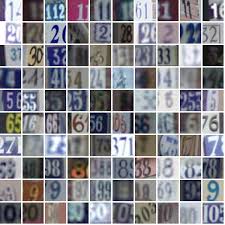}
    \caption{SVHN dataset}
    \label{fig:svhn}
\end{figure}

We normalized images in the dataset by making white the dominant color --- images with a majority of black pixels were inverted.
No augmentation was applied to the images.

\subsection{Software}

Our neural architecture is quite nonstandard, and the training algorithms are even more nonstandard.
Many machine learning frameworks can perform inferences on models like ours (the framework has to be flexible enough to allow scattered operations on tensors; in particular, TensorFlow can do this).
However, no present framework can \emph{train} such models.
Thus, we had to write the training and inference code in C++ from scratch.

This code --- an experimental machine learning framework with several examples --- can be downloaded from \url{https://www.alglib.net/strongnet/}.

\subsection{Network architecture}

\begin{figure}[htp]
    \centering
    \includegraphics[width=7cm]{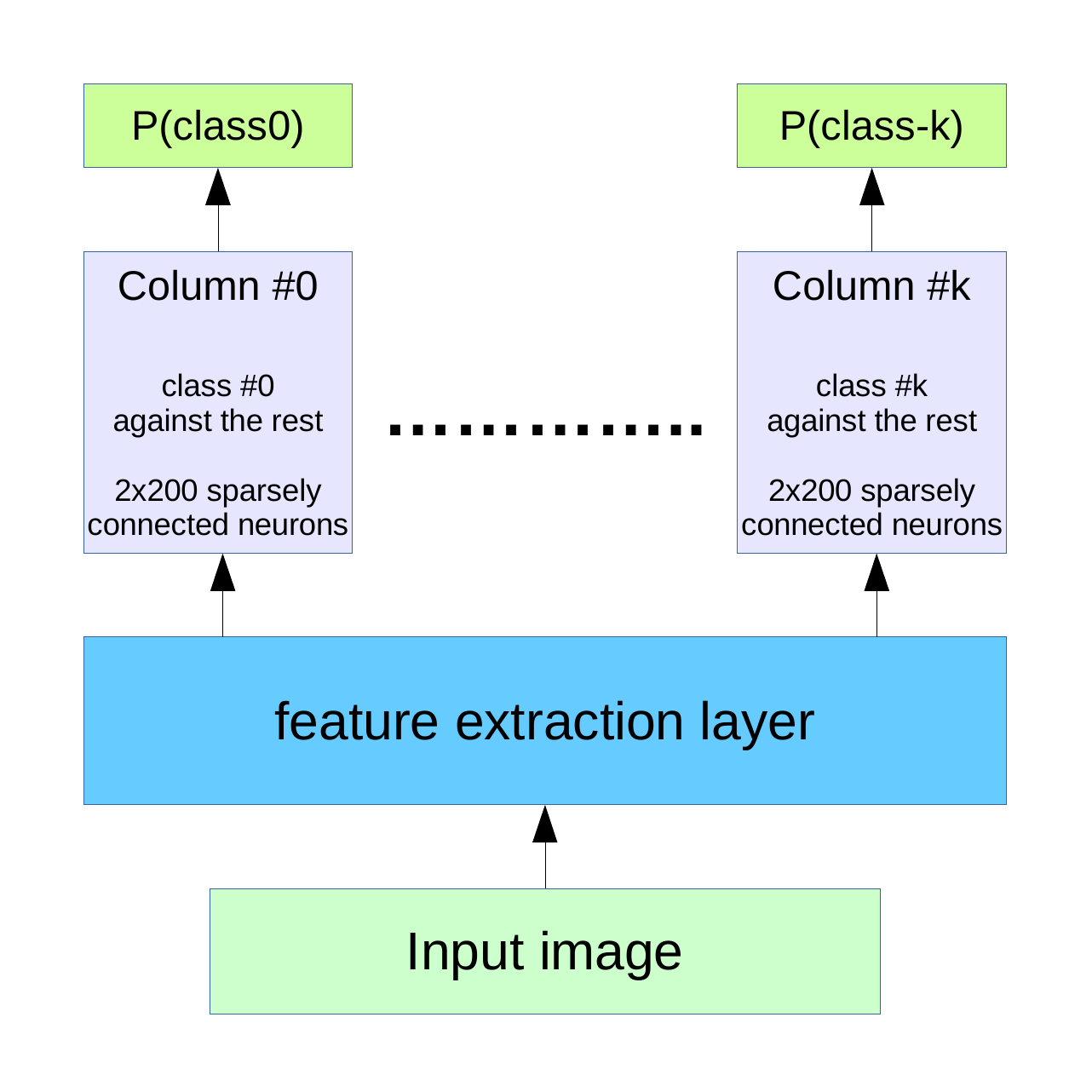}
    \caption{Network structure}
    \label{fig:resultsnetwork}
\end{figure}

In this work, we evaluated a multi-column architecture with a shared unsupervised feature detection layer and separate supervised classification columns (see Figure \ref{fig:resultsnetwork}).
The $K$-th column is individually trained to separate class $K$ from the rest of the dataset.

The feature detection layer has two separate blocks: contour (color-agnostic) features and color-based ones.
The contour filter bank has a capacity equal to 50 different filters.
These filters have a size of 6x6, which allows the detection of medium complexity shapes; that is, ones more complex than simple edges.
Each of these filters produces two features --- one corresponding to the "positive" phase and one to the "negative" phase --- so the total channel count is 100.
The color filter bank is much smaller and stores just 10 filters, each having a size of 4x4, which is adequate to detect uniformly colored patches.

In both cases (contour and color), we perform multiscale feature analysis, processing 32x32 (scale 0) and downsampled 16x16 (scale 1) versions of the image.
The contour block requires 4.6 MFLOP to be computed, while the color block needs 0.4 MFLOP.
Thus, the total amount of floating point operations required to perform initial feature detection is \textbf{5.0 MFLOP}.


Classification columns are composed of our novel strong neurons grouped into two sparsely connected "strong layers" followed by a single output sigmoid neuron (linear summator + logistic function).
Shortcut connections are present between all strong layers and outputs.

In our experiments, columns with widths equal to just 200 strong neurons were powerful enough to separate GTSRB classes.
Such columns needed roughly \textbf{0.007 MFLOP} (7000 FLOP).

The output of the $k$-th column is the probability of the image belonging to class $K$.
Due to logistic model properties, this probability is usually well calibrated.
However, it is important to remember that different columns are trained separately, so their outputs do not have to sum to one.

\subsection{Results: low-cost inference on GTSRB}

The GTSRB dataset has 43 classes, so our network has a shared feature detection layer and 43 class-specific sparse columns.
This means that the inference cost of our model is \textbf{$5.0+43\times0.007=5.3$ MFLOP}.
The test set error of our model on this dataset is \textbf{1.6\%}.

\begin{figure}[H]
    \centering
    \includegraphics[width=10cm]{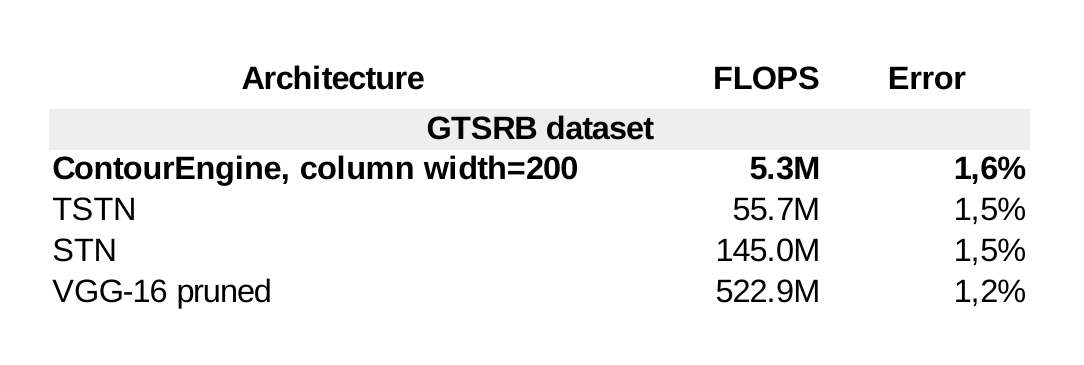}
    \caption{GTSRB: accuracy vs inference cost}
    \label{fig:gtsrbresults}
\end{figure}

The table above compares Contour Engine with Targeted Kernel Networks \cite{targetedkernelnets} and pruning \cite{yiming}.
Targeted Kernel Networks (TSTN and STN rows) reduce computational complexity by dropping some of the inner convolutions using attentional modulation.
They may be regarded as a type of spatial pruning.
The work by Yiming Hu et al. involved channel-based pruning performed using a genetic algorithm.
Contour Engine outperforms both approaches by an order of magnitude.

One more interesting point is that the $5.3$ MFLOP required by our model are mostly unsupervised.
Only $0.3$ MFLOP ($0.007$ MFLOP per class) are performed in the supervised part of our network.
Most of the time is spent on unsupervised preprocessing, which consumes about $95\%$ of the computational budget.
This result suggests that the actual complexity of the contour-based classification is on the kiloflop rather than on the megaflop or gigaflop scale.

\subsection{Results: low-cost inference on SVHN}

The Street View House Numbers dataset has 10 classes, so our network uses a shared feature detection layer similar to the one employed on GTSRB with 10 class-specific sparse columns.

We note here that in this task color does not carry any classification-related information (e.g., the green-vs-blue edge is important because it is an edge, not because it is green or blue), so we dropped the color part of the feature extraction layer.

The inference cost for our model was \textbf{4.8 MFLOP}, and the test set error was \textbf{4.8\%}.

\begin{figure}[H]
    \centering
    \includegraphics[width=10cm]{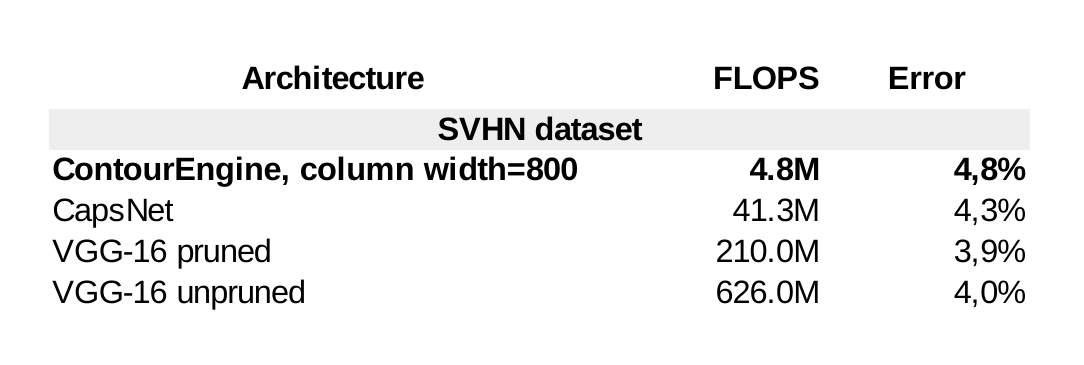}
    \caption{SVHN: accuracy vs inference cost}
    \label{fig:svhnresults}
\end{figure}

For this dataset, we compare our network with the pruning by Yiming Hu et al. (again) and with Capsule Networks (\cite{capsnets}, \cite{targetedkernelnets}).
Again, Contour Engine outperforms its competitors by an order of magnitude.

\subsection{Results: improved adversarial stability}

We tested the adversarial stability of the Contour Engine network trained on the SVHN dataset.
We used a powerful PGD attack (iterated FGSM with 20 iterations and backtracking line search) with the perturbation $L_\infty$-norm bounded by 0.01, 0.02 and 0.03.

\begin{figure}[H]
    \centering
    \includegraphics[width=10cm]{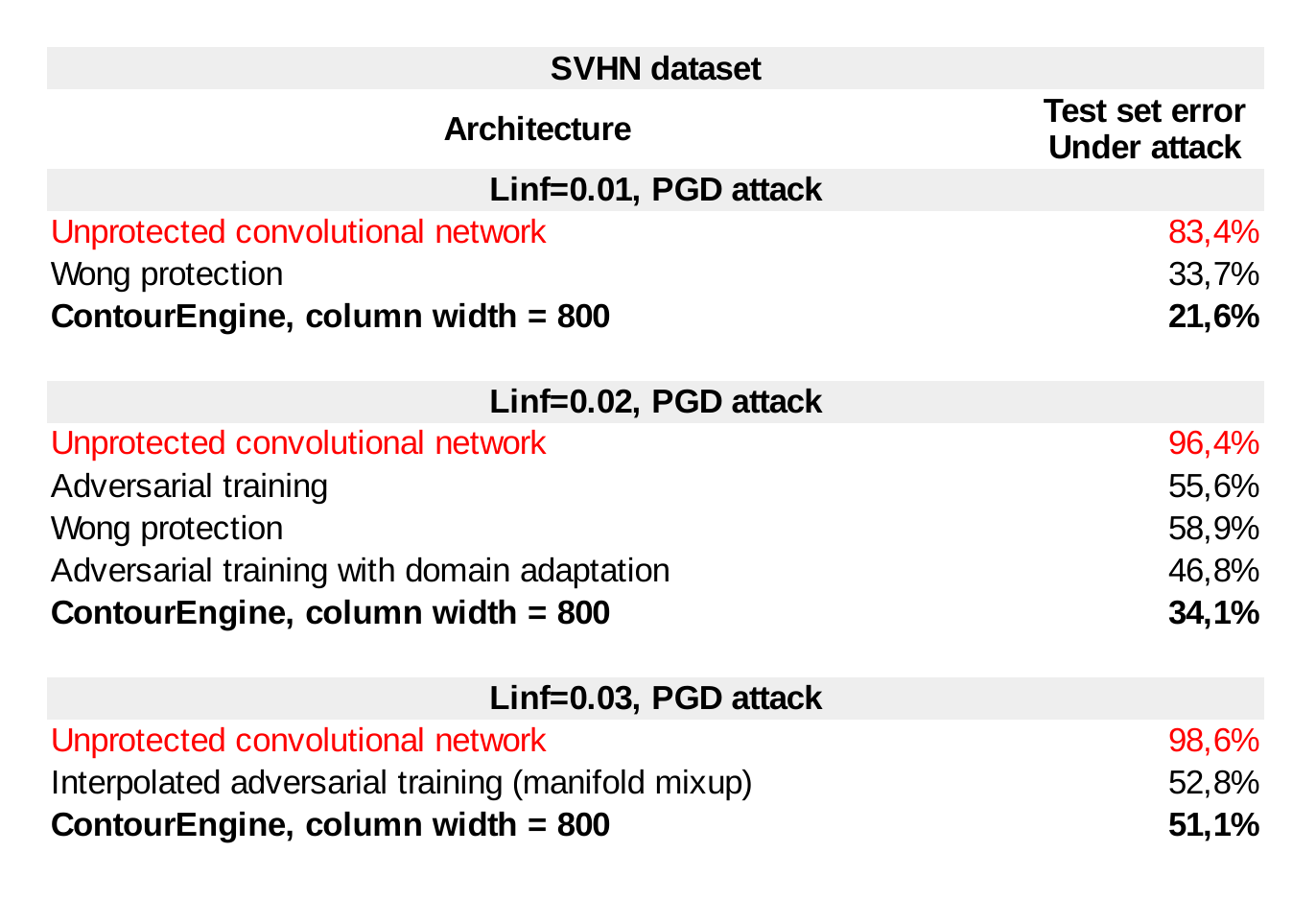}
    \caption{SVHN: adversarial attack success rate}
    \label{fig:adversarialresults}
\end{figure}

The table above compares the attack success rate for Contour Engine with reference values from three independent works (\cite{wongdefense}, \cite{atda}, \cite{iat}).
It can be seen that an unprotected network can be successfully attacked in 83\% cases with a perturbation as small as 0.01.
Different kinds of adversarial protection (when used on traditional summator-based networks) significantly reduce the attack success rate.
However, in all cases Contour Engine outperforms these results without any special counter-adversarial measures.

\subsection{Results: hardware requirements}

Our neural network has fairly low hardware requirements.
We already mentioned its low floating point count, but another interesting property is that it is easy to switch from floating point operations to fixed point ones.
Stability with respect to adversarial perturbations (maliciously targeted ones) implies stability with respect to perturbations arising from rounding (untargeted ones) --- thus one may expect graceful degradation with a progressive decrease in mantissa length.

Different parts of the network have different hardware requirements with respect to working accuracy:

\paragraph{Feature detection layer.} This part of the network is just a single layer of convolutions with bounded coefficients, performed on $[0,1]$-bounded inputs, producing $[0,1$-bounded outputs.
Thus, it can be efficiently implemented with no drop in the inference quality using just 8-bit fixed point inputs and outputs and 8-bit unsigned integer multiplicator/summator units with 24-bit accumulators.

\paragraph{Strong layers.} This part of the network can also be implemented with 8-bit fixed-point units.
With binary weights, this part of the network is multiplication free and summation free, so only 8-bit min and max units are needed.
With non-binary weights, strong neurons may need multiplication by fixed-point numbers with short mantissas (e.g., $1\nicefrac{1}{2}$), which may be performed with just a few shifts/adds.

\paragraph{Shallow classifier.} This part of network is just a single summator with bounded coefficients.
Hence, it may work well with 8-bit fixed point inputs and outputs, 8-bit unsigned integer multiplicator units and 24-bit internal accumulators.

In fact, our model's accuracy and stability results were obtained with 7-bit precision to store the activity matrices.
We had to utilize this reduced precision due to the immense memory requirements of some parts of our training algorithm. However, this also allowed us to experimentally verify our claims with low hardware requirements.
Experimenting with a 4-bit version of our network also looks promising.

\section{Summary}
\label{sect:conclusions}

In this work, we have proposed a novel model of the artificial neuron --- the strong neuron --- which can separate classes with decision boundaries more complex than hyperplanes and which is resistant to adversarial perturbations of its inputs.
We proved that our proposal is a fundamental and well-motivated change and that constituent elements of our strong neuron, $min$/$max$ units, are the only robust implementations of the AND/OR logic.
We also proposed a novel training algorithm that can generate sparse networks with $O(1)$ connections per strong neuron, a result that far surpasses any present advances in neural network sparsification.

State-of-the-art efficiency (inference cost) is achieved on GTSRB and SVHN benchmarks.
We also achieved state-of-the-art results in terms of stability against adversarial attacks on SVHN --- without any kind of adversarial training --- which surpassed much more sophisticated defenses.
Further, our network has low hardware requirements and gracefully degrades when numerical precision is decreased (we managed to achieve the results listed above using just 8-bit fixed point math for the unit activities).

One more interesting result is related to our decision to separate unsupervised feature detection and supervised classification.
We found that Contour Engine spends most of the inference time in the unsupervised preprocessor --- less than 10.000 FLOP per class is used by the supervised part of the network (one which is composed of strong neurons). 
This result suggests that contour recognition is much easier than was previously thought. Once initial unsupervised image preprocessing is done, centered contours can be recognized with just a few kiloflops.

Finally, we want to highlight future directions of our work:

\begin{itemize}
\item \textbf{Convolutional training.}
Our proof-of-concept network is nonconvolutional, which limits its applicability to well-centered image recognition problems, such as MNIST, GTSRB, and SVHN.
The next step is to implement computationally feasible convolutional training.
\item \textbf{Better adversarial stability.}
We already achieved state-of-the-art stability with a simple linear output.
However, we believe that further improvements are possible with a better shallow classifier layer (output layer).
This layer is the only adversarially unstable part of the network --- we managed to reduce the problem of building a \emph{deep} and robust network to one of building a \emph{shallow } and robust one.
One promising robust classifier model is a maxout\cite{maxout} neuron with an $L_1$ constraint on internal linear subunits.
\item \textbf{Transfer learning and fast retraining.}
The filters of the unsupervised feature detection layer look quite generic (edges, bars, blobs, arcs), which strongly suggests that this layer could be reused across multiple pattern detection problems.
Thus, one obvious direction of research involves the transfer properties of the feature detection layer.
Furthermore, we feel that the strong neurons generated by the sparse training algorithm may also allow some limited reuse.
When combined with extremely cheap inference performed by strong neurons, this opens the door to pretrained "universal column," which contain strong neurons capable of detecting a wide range of "popular contours."
\end{itemize}

\bibliographystyle{alpha}

\newpage

\section{Appendix A: proofs of theorems 1 and 2}
\label{sect:appendixa}

\paragraph{Theorem 1: $L_\infty$-nonexpansive AND problem.}
$\exists!{\enspace}f(x,y)=min(x,y)$ such that following holds:
\begin{enumerate}
\item[C1] $f(x,y)$ is defined for $x,y \in [0,1]$
\item[C2] $f(0,0)=f(0,1)=f(1,0)=0$
\item[C3] $f(1,1)=1$
\item[C4] $a{\leq}A,\ \ b{\leq}B \implies f(a,b){\leq}f(A,B)$ (monotonicity)
\item[C5] $|f(a+{\Delta}a,b+{\Delta}b)-f(a,b)| \leq max(|{\Delta}a|,|{\Delta}b|)$
\end{enumerate}

\paragraph{Proof.} We will prove Theorem 1 by demonstrating that conditions C1...C5 constrain $f(x,y)$ in such a way that the only possible solution is $f(x,y)=min(x,y)$.

The monotonicity condition C4 combined with C2 means that

\begin{equation}  \label{eq:f0y}
{\forall}\ y{\in}[0,1]\ \ \ f(0,y)=0
\end{equation}

Condition C5, when combined with C2 and C3, means that ${\forall}y{\in}[0,1]\ \ f(y,y)=y$. Indeed, C5 combined with C2 means that $|f(y,y)-f(0,0)| \leq |y|\ \implies\ f(y,y){\leq}y$.
Similarly, C5 combined with C3 means that $|f(y,y)-f(1,1)| \leq |1-y|\ \implies\ f(y,y){\geq}y$.
As result, we have

\begin{equation}  \label{eq:fyy}
{\forall}\ y{\in}[0,1]\ \ \ f(y,y)=y
\end{equation}

Similarly to the previous paragraph, condition C5 combined with \ref{eq:f0y} and \ref{eq:fyy} constrains function values between $f(0,y)$ and $f(y,y)$ to

\begin{align*}
{\forall}\ 0{\leq}x{\leq}y{\leq}1\ \ \ f(x,y)=x=min(x,y)
\end{align*}

Due to the symmetry of the problem, it is obvious that the following also holds:

\begin{align*}
{\forall}\ 0{\leq}y{\leq}x{\leq}1\ \ \ f(x,y)=y=min(x,y)
\end{align*}

So, finally,

\begin{align*}
{\forall}x,y\in[0,1]\ \ \ f(x,y)=min(x,y)
\end{align*}

which has been shown.

\paragraph{Theorem 2: $L_\infty$-nonexpansive OR problem.}
$\exists!{\enspace}g(x,y)=max(x,y)$ such that following holds:
\begin{enumerate}
\item[C1] $g(x,y)$ is defined for $x,y \in [0,1]$
\item[C2] $g(0,0)=0$
\item[C3] $g(0,1)=g(1,0)=g(1,1)=1$
\item[C4] $a{\leq}A,\ \ b{\leq}B \implies g(a,b){\leq}g(A,B)$ (monotonicity)
\item[C5] $|g(a+{\Delta}a,b+{\Delta}b)-g(a,b)| \leq max(|{\Delta}a|,|{\Delta}b|)$
\end{enumerate}

\paragraph{Proof.} Similarly to the previous proof, we will prove Theorem 2 by demonstrating that conditions C1...C5 constrain $g(x,y)$ in such a way that the only possible solution is $g(x,y)=max(x,y)$.

C5 combined with C2 and C3 constrains $g(x,y)$ along $x=y$: $g(0,0)=0 \implies g(y,y) \leq y$ and $g(1,1)=1 \implies g(y,y) \geq y$, so finally we have

\begin{equation}  \label{eq:gyy}
\forall\ y\in[0,1]\ \ \ g(y,y)=y
\end{equation}

Similarly, for $g(0,y)$ from the nonexpansivity constraint C5 combined with boundary values $g(0,0)=0$ and $g(0,1)=1$, it immediately follows that

\begin{equation}  \label{eq:g0y}
\forall\ y\in[0,1]\ \ \ g(0,y)=y
\end{equation}

and, due to monotonicity constraint C4, from \ref{eq:gyy} and \ref{eq:g0y} we get

\begin{align*}
\forall\ 0 \leq x \leq y \leq 1\ \ \ g(x,y)=y=max(x,y)
\end{align*}

Due to the obvious symmetry, it is easy to prove that

\begin{align*}
\forall\ x,y\in[0,1]\ \ \ g(x,y)=max(x,y)
\end{align*}

which has been shown.

\section{Appendix B. The feature detection layer}
\label{sect:appendixb}

In this section we discuss a feature detection layer based on \cite{Coates11} with several proposed improvements.
There are several interesting ideas we want to share here, so this section is quite long.
Nevertheless, we deem this part of our work as less important than the results on strong neurons, so we moved it to the end of the article.

Modern convolutional networks tend to have many layers with filters as small as 3x3.
One well-known pattern is to have two layers with 3x3 convolutions followed by a max-pooling layer.
Almost all architectures lack a clear distinction between feature extraction and subsequent geometric inference --- both tasks are performed using the same sequence of standard building blocks.
Due to the quadratic dependence between the network width and weights count, preference is given to deep and narrow networks --- making the network 2x deeper and 2x narrower results in a 2x decrease in computing power.

In contrast, our neural architecture has sparse layers with $O(1)$ connections per neuron.
It thus inherently favors shallow and wide networks.
Another difference from traditional architectures is that our strong neurons can perform logical inferences on low-level features, although they cannot \emph{produce} these features from raw pixel values.
Thus, a separate feature extraction block is essential in order to "prime" Contour Engine.

The purpose of our feature extraction layer is to describe an input image using a rich dictionary of visual words.
The description includes features such as oriented edges, more complex shapes, colors and gradients, computed at multiple scales and orientations.

The following subsections discuss our implementation of the feature extraction layer, starting from the very basic setup and progressively improving it.

\subsection{The basic structure}

The basic implementation of the feature extraction unit is a single layer of 4x4 and/or 6x6 convolutions followed by normalization and sparsification (see \cite{Coates11}) layers:

\begin{align*} 
y_{raw}[i,j,k] &= ReLU\left(CONV(W,x)\right) \\ 
y_{sparse}[i,j,k] &= ReLU\left(y_{raw}[i,j,k] - \lambda\underset{k}{MEAN}(y_{raw}[i,j,k])\right) \\
y_{nrm}[i,j,k] &= \frac{y_{sparse}[i,j,k]}{\epsilon+\max\limits_{i,j,k} y_{sparse}[i,j,k]}
\end{align*}

where $W$ is a $Kx3xMxM$ tensor (here $K$ is an output filter count, $M$ is a convolution size and $3$ stands for RGB input) and $\lambda$ is a tunable sparsification parameter.
The typical amount of filters within feature banks ranges from 8 (just edge detectors) to 100 (medium-complexity shapes) features.

We experimented with different methods of generating feature banks and found that training them in the completely unsupervised manner (see \cite{Coates11}) tends to give good results with interesting generalization properties, which will be discussed later.

\subsection{Separating contour and color}

One improvement we propose is to separate contour-based and color-based features.
We require the former to be color-agnostic (the feature detector output does not change under permutation of RGB channels) and the latter to be lightness-agnostic (the feature detector output does not change with the addition/subtraction of gray color).

We have several reasons behind our proposal.
First, it is well known that the human visual cortex (the best universal visual processor known so far) performs separate processing of contour and color signals in the first regions of the ventral stream, also known as the "what pathway."
We want to duplicate it here because our work was partially inspired by unique properties of the human visual system.
Second, having such orthogonality in our model accelerates training in later stages (creating sparse connectivity) because it greatly reduces the number of possible connections in the network.
Finally, such separation makes our network more controllable --- we can easily measure the amount of information provided by the edges and color and easily introduce some invariants into the model (e.g., invariance with respect to various color and lightness corrections).

Color-agnostic processing can be implemented by requiring that components of the tensor $W$ corresponding to different RGB channels have the same value.
However, we prefer to explicitly replace the $Kx3xMxM$ weight tensor $W$ with the $KxMxM$ tensor $W_L$:

\begin{math}
y_{L,raw}[i,j,k] = ReLU\left(CONV(W_L,\frac{1}{3}\left(x_R+x_G+x_B\right))\right)
\end{math}

One more normalization we introduce is a requirement that the feature detector output be invariant with respect to lightness shift (addition/removal of the gray color).
Mathematically, this condition means that we require tensor elements within each filter to sum to zero:

\begin{math}
{\forall}k:\quad \sum\limits_{i,j}W_L[k,i,j] = 0
\end{math}

One possible way to enforce such requirements is to tweak the data fed to "k-means over image patches" procedure proposed by Coates et al.
Color-agnostic filters can be learned by replacing colors with monochrome values prior to running k-means.
The second requirement --- the invariance wrt lightness shift --- can be enforced by substracting the mean lightness from image patches.

Similarly, color-based lightness-agnostic processing can be implemented by requiring that components of the weight tensor $W$ corresponding to different RGB channels sum to zero (invariance wrt to lightness shift is implicitly enforced by this constraint):

\begin{math}
\forall i,j,k:\quad W_C[k,0,i,j]+W_C[k,1,i,j]+W_C[k,2,i,j] = 0
\end{math}

As with color-agnostic filters, color-based ones can be learned through manipulation with data sent following the Coates procedure --- one can simply subtract the lightness value from each pixel.

The following filters were learned by running this procedure on the CIFAR dataset:

\begin{figure}[H]
    \centering
    \includegraphics[width=10cm]{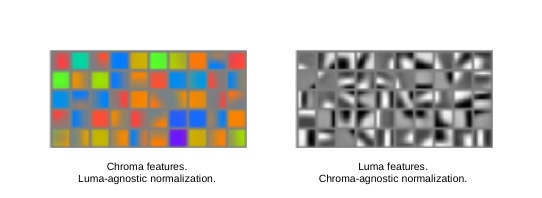}
    \caption{Chroma and luma filters}
    \label{fig:appbfilters}
\end{figure}

\subsection{Downsampling (max-pooling) layer}

The max-pooling layer is well known for its ability to simultaneously reduce the dimensionality of the data and improve its linear separability (the latter is achieved due to the introduction of shift-invariance).
We again refer to \cite{Coates11} for some interesting quantative results.
In this section, we focus on the max-pooling layer, which performs max-downsampling of the input tensor (pooling with a filter width equal to the stride).
The question is, what downsampling factor is the best one?

Numerical experiments showed that, for 4x4- and 6x6-sized features, good results could be achieved with 2x downsampling.
This provides a good balance between generalization and loss of essential spatial information.
While 4x downsampling loses too much information to be used alone, it can supplement 2x-downsampled activities if both are used together.

\subsection{Feature detection at multiple scales}

Although the initial formulation covers just small 4x4 or 6x6 image patches, one may reasonably want to have a multiscale description that includes both small (e.g., ~4x4 pixels), medium (~8x8) and large (~16x16) features.

Traditional convolutional architectures do not explicitly form such multiscale representations.
Since the beginning, the dominant approach has been to stack standard building blocks and allow SGD to do the rest.
We, however, aim to develop an architecture that performs some standardized kinds of processing (feature extraction, spatial pooling, multiscale processing) in the standardized manner with a limited amount of controllable nonlinearities learned.

\subsection{Introducing completeness}

Now, we have everything we need to prime Contour Engine --- shape/color separation, multiple downsampling levels and multiscale image processing.
The key parts of our feature detection layer are present.
However, we may add one more improvement --- completeness.
It is preferable to have a feature detection layer that is complete under some particular set of transformations.

For example, if feature $F_0$ detects some particularly oriented shape, the feature detection layer may also be required to have $F_1$, $F_2$ and $F_3$ that detect the same shape rotated by $90{\degree}$, $180{\degree}$ and $270{\degree}$,respectively.
Another option is to require completeness with respect to permutations in color space --- one may require a color gradient to be detected for any combination of constituent colors (red-green, red-blue, green-blue, yellow-blue, violet-green and so on).

This requirement may be a bit too much for specialized computer vision systems like those that detect traffic lights --- red blobs against black backgrounds are important, but violet blobs against green background are irrelevant for solving the problem.
However, to design a general purpose vision system that can be specialized for any task, having such a feature detection layer may be essential for success.

\emph{What is usually achieved by training a "prototype network" on a large, diverse dataset (say, ImageNet) can also be achieved by introducing completeness in a network trained on a much smaller dataset}.

In this work, however, we focus on another aspect of complete feature subsets: computational complexity.
Some types of completeness allow us to achieve constant 2x-6x performance boost, that is, to have subsets of two features (completeness with respect to lightness inversion) or six features (completeness with respect to color rotation) computed in roughly the same time as is usually needed to compute just one feature.

Completeness with respect to lightness inversion means that color-agnostic features now come in two subsets --- corresponding to the "positive phase" of some filter and corresponding to the "negative phase":

\begin{align*} 
y_{f}[i,j,k] &= CONV(W,x) \\
y_{raw}[i,j,k] &= CONCAT\left[ ReLU(+y_f), ReLU(-y_f) \right] \\ 
y_{sparse}[i,j,k] &= ReLU\left(y_{raw}[i,j,k] - \lambda\underset{k}{MEAN}(y_{raw}[i,j,k])\right) \\
y_{nrm}[i,j,k] &= \frac{y_{sparse}[i,j,k]}{\epsilon+\max\limits_{i,j,k} y_{sparse}[i,j,k]}
\end{align*}

This improvement allows us to achieve a constant 2x performance boost for the color-agnostic part of our feature detection layer.
This means that we either can have a 2x wider layer (more features detected) with the same performance budget, or alternatively, we can have roughly the same level of quality with a 2x smaller running time.
Similar, albeit more complex changes can be made to introduce completeness with respect to rotations in color space.

Capturing both positive and negative phases of ReLU units was proposed long before this work (e.g., \cite{crelu}, \cite{maxmin}).
However, most previous authors failed to consider the fact that capturing positive/phases is just a special case of the more general movement toward having a complete feature detection layer.

\end{document}